\def\year{2019}\relax
\documentclass[letterpaper]{article} 
\usepackage{aaai19}  
\usepackage{times}  
\usepackage{helvet}  
\usepackage{courier}  
\usepackage{url}  
\usepackage{graphicx}  
\usepackage{natbib}

\usepackage[T1]{fontenc}    
\usepackage[utf8]{inputenc} 
\usepackage[draft]{hyperref}
\usepackage{hypernat}
\usepackage{booktabs} 
\usepackage{amsfonts}       
\usepackage{nicefrac}

\usepackage{microtype} 
\usepackage{smartref}
\usepackage{amsthm}
\usepackage{tikz}
\usepackage{float}
\usepackage{subcaption}
\usepackage{stackengine}
\usepackage{amsmath} 
\usetikzlibrary{positioning}
\usetikzlibrary{shapes.geometric, arrows}
\usepackage{multirow}
\usepackage{enumitem}
\usepackage{algorithm}
\usepackage{algorithmic}
\usepackage[justification=centering]{caption}
\usepackage{todonotes}

\usepackage{amssymb} 
\usepackage{amsthm}
\usepackage{epstopdf} 
\usepackage{float}

\newcommand{\cR}{\mathcal{R}}

\newcommand{\Nat}{\mathbb{N}}

\newcommand{\Esp}{\mathbb{E}}

\newcommand{\argmax}{\mathop{\mathrm{argmax}}}

\definecolor{myred}{RGB}{219, 48, 122}

\newtheorem{remark}{Remark}

\begin{document}
 \setcounter{secnumdepth}{2}
%
\title{On-line Adaptative Curriculum Learning for GANs}

\author{Thang Doan\textsuperscript{1,8}, Jo\~ao Monteiro\textsuperscript{2}, Isabela Albuquerque \textsuperscript{2}, Bogdan Mazoure \textsuperscript{3},\\
{\bf \Large  Audrey Durand\textsuperscript{4,5}, Joelle Pineau\textsuperscript{4,5,6}, R Devon Hjelm\textsuperscript{5,7}} 
\\\textsuperscript{1}Desautels Faculty of Management, McGill University\\\textsuperscript{2}INRS-EMT, Universit\'e du Qu\'ebec\\\textsuperscript{3}Department of Mathematics \& Statistics, McGill University\\\textsuperscript{4}School of Computer Science, McGill University\\\textsuperscript{5}Mila -- Quebec Artificial Intelligence Institute\\\textsuperscript{6}Facebook AI Research,\textsuperscript{7}Microsoft Research Montreal}


\maketitle
\begin{abstract}
Generative Adversarial Networks (GANs) can successfully approximate a probability distribution and produce realistic samples. However, open questions such as sufficient convergence conditions and mode collapse still persist. In this paper, we build on existing work in the area by proposing a novel framework for training the generator against an ensemble of discriminator networks, which can be seen as a one-student/multiple-teachers setting. We formalize this problem within the full-information adversarial bandit framework, where we evaluate the capability of an algorithm to select mixtures of discriminators for providing the generator with feedback during learning. To this end, we propose a reward function which reflects the progress made by the generator and dynamically update the mixture weights allocated to each discriminator. We also draw connections between our algorithm and stochastic optimization methods and then show that existing approaches using multiple discriminators in literature can be recovered from our framework. We argue that less expressive discriminators are smoother and have a general coarse grained view of the modes map, which enforces the generator to cover a wide portion of the data distribution support. On the other hand, highly expressive discriminators ensure samples quality. Finally, experimental results show that our approach improves samples quality and diversity over existing baselines by effectively learning a curriculum. These results also support the claim that weaker discriminators have higher entropy improving modes coverage.





\end{abstract}

\section{Introduction}
Generative Adversarial Networks~\citep[GANs, ][]{originalGAN} have reshaped the state of machine learning in tasks that involve generating data. 
A GAN is an unsupervised method that consists of two neural networks, a generator and a discriminator, with opposing (or \emph{adversarial}) objectives. The typical goal of the generator is to transform noise (e.g., drawn from a normal distribution) into samples whose statistical and structural characteristics match well those of an empirical target dataset (such as a collection of images).
The discriminator, which acts as an \emph{adversary} to the generator, needs to discriminate between (or \emph{classify}) samples as coming from the real data or the generator.

While GANs can achieve impressive qualitative performance~\citep[most notably with image data, e.g., see][]{stabilizing_GAN_roth,miyato2018spectral,GrowingGAN}, the most successful methods depart from the original formulation to address various instabilities and other optimization difficulties~\citep{arjovsky2017towards,WGAN}. 
One such difficulty in training GANs occurs when the generator produces samples only from a small subset of the target distribution, a phenomenon known as \emph{missing modes}~\citep[a.k.a., \emph{mode-dropping}, e.g. see][]{che2016mode}.
Numerous works try to address the problem by modifying the original objective, such as unrolling~\citep{UnrolledGAN}, aggregating samples~\citep{PacGAN}, stacked architectures~\citep{stackedGAN,GrowingGAN}, mutual information / entopy maximization~\citep{MINE}, multiple discriminators~\citep{RandomProjectionGAN,GoGANs}, or multiple generators~\citep{tolstikhin2017adagan,MGAN,GANpartbypart}.

\begin{figure}
    \hspace*{-0.5cm}  
    \centering
    \includegraphics[width=6cm,height=5cm]{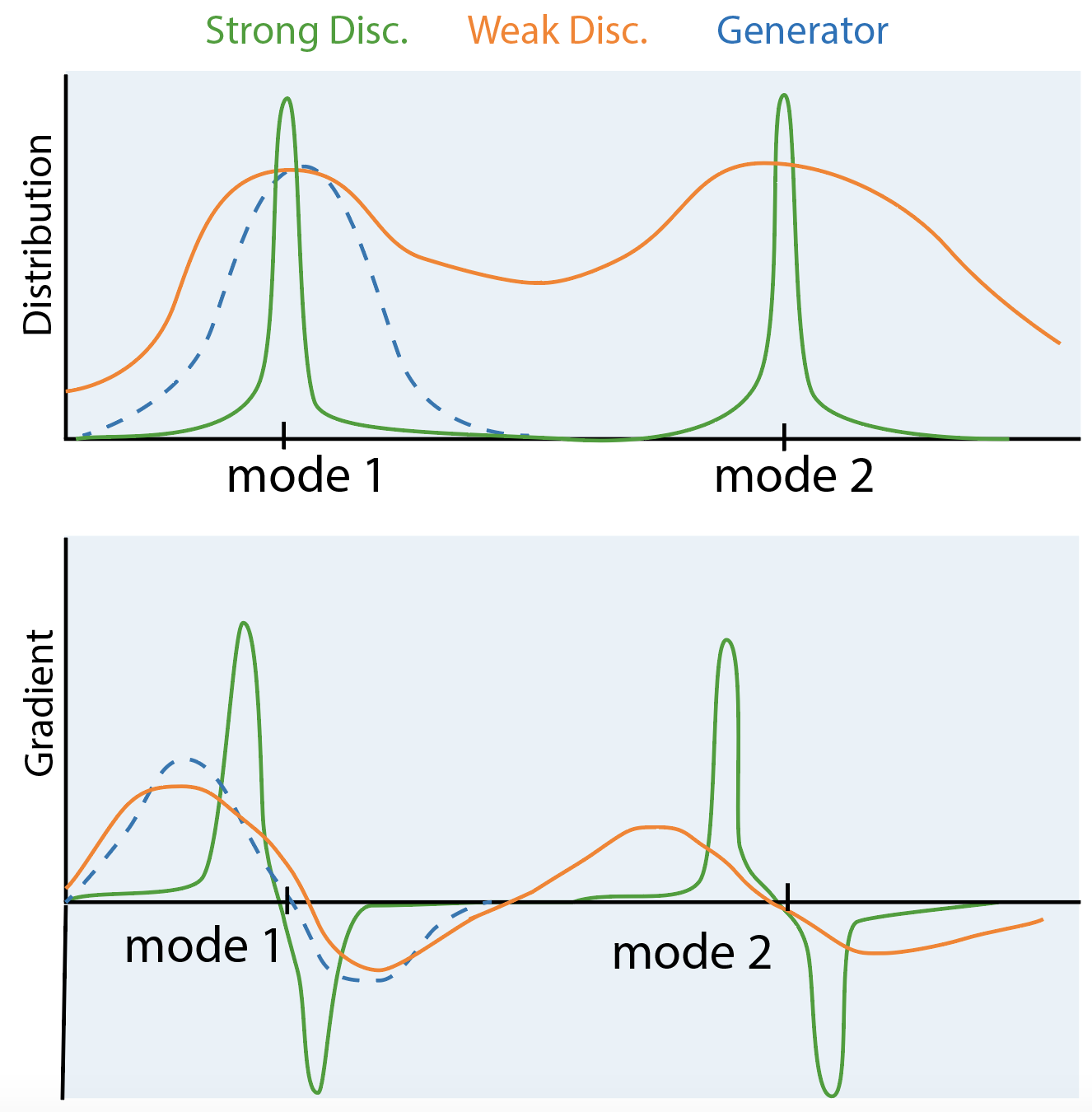}
    \caption{Recovering dropped modes via multiple discriminators. The weak discriminator provides feedback, allowing the generator to recover forgotten modes. The strong discriminator experiences vanishing gradient and cannot help the generator to recover modes.}
    \label{fig:cartoon}
\end{figure}

In our work, we follow the intuition that missing modes in GANs are due in part to mode-specific vanishing gradients.
As a simple illustrative example which we explore in detail in our experiments below (Fig.~\ref{fig:cartoon}), consider a discriminator that is well representing the target distribution and a generator that is only generating a subset of the modes in the data.
If any of the missing modes are \emph{disjoint} from those represented in the generator (i.e., are composed of sets of features with low intersection), there is no way for the generator to receive gradient signal on missing modes from the discriminator.
However, if the discriminator only represents the data approximately (in the sense that it also cannot fully distinguish between these modes), it may be possible to recover the missing mode gradient signal. If this can be achieved by using a low capacity\footnote{Throughout the paper, we refer to \emph{capacity} as the architecture size of a given neural network in terms of number of parameters.} discriminator, it is ultimately undesirable given that the end goal is to generate samples that resemble well the target dataset. From now on, we will refer to such low capacity discriminators as \emph{weak} and to high capacity discriminators as \emph{strong}. In order to ensure both high quality and mode coverage, we consider multiple discriminators~\citep[as in][]{GMAN} with different strengths to train the generator. We propose to train the generator using a curriculum based on an on-line multi-armed bandit algorithm~\citep{teacher_student, AutomatedCuri}, dynamically changing the weight/resources allocated to each discriminator, which we show is crucial for achieving good results.
Our primary contributions are:
\vskip 0.1in
\begin{enumerate}[leftmargin=*,noitemsep,nolistsep]
\item We provide important insights into the missing mode problem as demonstrated by the gradient signal available to the generator from the discriminator.
\item As a potential solution to the missing modes problem, we introduce a new framework based on adversarial bandits~\citep{Littlestone1994,Auer1995,Freund1997} resource allocation, where the generator gets its training signal from a set of teacher networks with increasing capacity.
\item We show that the proposed approach leads to a curriculum learning characterized by successive phases of the generator prioritizing different discriminators.
\end{enumerate}

\vskip 0.1in
The remainder of this paper is organized as follows. Previous literature relevant to this work is briefly reviewed on Section~\ref{sec:rel_work}. The proposed approach is formally introduced in Section~\ref{sec:acgan}, and an empirical analysis is reported in Section~\ref{sec:experiments}. Conclusions and future directions are finally presented in Section~\ref{sec:conclusion}.



\section{Related Work}
\label{sec:rel_work}
\subsubsection{Mode coverage and data / model augmentation} The intuition that missing modes are due to vanishing gradients resonates with some successful approaches on stabilizing and improving GAN training through data and model augmentation.
Instance noise~\citep{arjovsky2017towards} has been shown to improve stability~\citep[see also][]{stabilizing_GAN_roth}, which can be understood as smoothing the data modes in the pixel space.
Progressively reducing the downsampling through training (either by copying parameters or feeding low resolution samples into a larger generator) have also been considered previously~\citep{stackedGAN,GrowingGAN} as solutions to increase mode overlap.
This is akin to a hand-crafted curriculum, progressively increasing the difficulty of the problem at a-priori chosen points in the complete training procedure.

\subsubsection{Multiple discriminators and generators} Several works have also incorporated multiple generators or discriminators in order to improve learning. 
Multiple-generator methods~\citep{tolstikhin2017adagan,MGAN,GANpartbypart} typically work by encouraging the generators to divide the task of generating by modes in the target dataset (without additional supervision). 
Using multiple discriminators~\citep{RandomProjectionGAN,GoGANs}, on the other hand, is known to provide a better learning signal for the generator if said discriminators compositionally represent well the target datasets.
Closest to our work, \citet{GMAN} consider discriminators of different complexity to provide varied signal. We will show that wisely designing the reward allows to track the progress made by the generator and encourages a curriculum learning.




\subsubsection{Multi-armed bandit as a curriculum learning method for GANs} 
Curriculum learning~\citep{CL} phrases a given machine learning problem as a set of tasks of increasing difficulty. 
GANs can also be said to share aspects with curriculum learning: the discriminator defines an objective of progressive difficulty, 

thus allowing the generator to gradually learn to more faithfully mimic the target distribution. However, there is no explicit mechanism to encourage a sensible curriculum for either model. 
For example, if the discriminator learns to represent disjoint modes faster than the generator learns to cover them, this can lead to the generator missing modes with no gradient signal to recover. 

In this paper, we propose an algorithm which gives rise to a curriculum in a direct manner.
Our approach borrows from curriculum learning in multi-armed bandit setting~\citep{teacher_student,AutomatedCuri}, where learning is typically done by measuring the change in a performance criterion of a given agent (i.e. a loss function, score or gradient norm can be used) that appears to affect the form of the optimal policy. 
In our method, given a set of discriminators, the goal is to weight the feedback received by the generator proportionally to the information contained in the gradients from each discriminator.





\section{Adaptative Curriculum GAN}
\label{sec:acgan}
Here we formulate the problem and approach for training a single generator on a target dataset using a curriculum over multiple discriminators, which we call \emph{Adaptative Curriculum GAN} (acGAN).
First, define a generator function, $G: \mathcal{Z} \mapsto \mathcal{X}$, which maps  noise from a domain $\mathcal{Z}$ to the domain of a target dataset, $\mathcal{X}$ (such as the space of images).
Let $p(x)$ denote the target density~\footnote{Here, we assume for the sake of notation that the target data admits a density.}, and let $p(z)$ denote the prior density defined on $Z$ used to draw noise samples for input into the generator. 
We wish to train this generator function using $N$ discriminators, $\mathcal{D} = \{D_i : \mathcal{X} \mapsto \mathbb{R}\}_{i=1}^N$, such that on each episode $t$, we select the mixture of discriminators that provides the best learning signal.

\subsection{Mixing discriminators}

This mixture-of-experts problem, where each discriminator plays the role of a teacher, can be tackled under the full-information adversarial bandit setting~\citep{Littlestone1994,Freund1997,Auer1995}.
On each episode $t$, a bandit player associates normalized weights $\Pi(t) = \{\pi_i(t)\}_{i=1}^N$ with discriminators $\{D_i\}_{i=1}^N$. The generator is then trained based on the mixture described by $\Pi(t)$, and a reward $\mathcal{R}_i(t)$ is observed for each discriminator $D_i$, characterizing the generator's improvement with respect to $D_i$.
Let $\cR(t) = \sum_{i=1}^{N}\pi_{i}\cR_{i}(t)$ denote the total observed reward at time $t$.
The goal of the player is to learn the optimal policy
$\Pi^\star(t) := \argmax_{\Pi \in \Delta(N-1)} \Esp_{\Pi(t), p(z)}[\mathcal{R}(t)]$ that maximizes the expected total reward\footnote{$\Delta(N-1)$ denotes the standard simplex on $\mathbb{R}^N$.}.

The Hedge algorithm~\citep{Freund1997}, also known as Boltzmann or Gibbs distribution, addresses this full-information game by maintaining probabilities
\begin{align}
\pi_{i}(t)=\frac{\exp{\lambda Q_i(t)}}{ \sum\limits_{j=1}^N \exp{\lambda Q_j(t)} }, \quad \lambda \geq 0,
\label{eq:boltzman_policy}
\end{align}
for each discriminator $D_i$, where $Q_i(t)$ estimates the gain of $D_i$ at episode~$t$. In this case, $\lambda$ is a parameter of the distribution: $\lambda=0$ corresponds to a uniform distribution over all models.
We found experimentally that using a moving average on previous rewards~\citep[which also featured in][]{teacher_student} stabilizes the training:
\begin{align}
Q_i(t)=\alpha \cR_{i}(t) + (1 - \alpha) Q_i(t-1),
\label{eq:MA_Q_value}
\end{align}
where $\alpha \in (0,1)$ is the smoothing parameter.
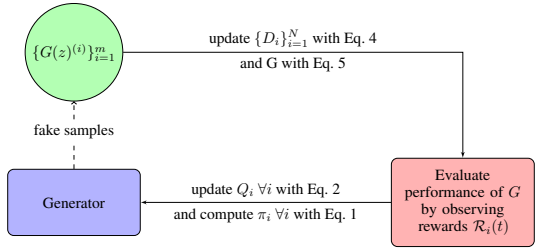
\begin{figure}
\hspace*{-0.4cm}
\centering
\resizebox{0.40\textwidth}{!}{
\begin{tikzpicture}[node distance=0.8cm][transform canvas={scale=0.5}]

\tikzstyle{Multi-Armed Bandits} = [rectangle,rounded corners, minimum width=3cm, minimum height=2cm, text centered,text width=3cm, draw=black, fill=red!30]
\tikzstyle{Generator} = [rectangle, rounded corners, minimum width=3cm, minimum height=1.5cm, text centered, draw=black, fill=blue!30]
\tikzstyle{fake_sample} = [circle, minimum width=1.2cm, minimum height=1.2cm,text centered, draw=black, fill=green!30]
\tikzstyle{MAB_pulling} = [rectangle, rounded corners, minimum width=2cm, minimum height=1cm,text centered,text width=2.5cm, draw=black, fill=orange!30]
\tikzstyle{arrow} = [draw, -latex']

\node (main) [Generator]                                                    {Generator};
\node (gg)  [above of=main,yshift=0.8cm]                                    {fake samples};
\node (G_fake) [fake_sample, above of=gg,yshift=1.0cm]                       {$\{G(z)^{(i)}\}_{i=1}^{m}$};
\node  (MAB) [Multi-Armed Bandits,right of=main, xshift=8cm,yshift=0.0cm]      {Evaluate \\ performance of $G$  \\ by observing rewards $\mathcal{R}_{i}(t)$};

\draw[dashed] (main) -- (gg);
\draw[dashed,->]  (gg.north) --  (G_fake.south);


\draw[arrow] (G_fake)  -|  (MAB.north) node [above,pos=0.25] {update $\{D_{i} \}_{i=1}^{N}$ with Eq.~\ref{eq:gan_disc_loss} } node [below,pos=0.25] {and G with Eq.~\ref{eq:gan_gen_loss} }; 
\draw[arrow] (MAB.west)  --  (main.east) node [above,pos=0.5] {update $Q_i~\forall i$  with Eq.~\ref{eq:MA_Q_value}} node [below,pos=0.5] { and compute $\pi_i~\forall i$ with Eq.~\ref{eq:boltzman_policy}} ; 
\end{tikzpicture}
}
\caption{Proposed procedure for training the generator}
\label{fig:tikz_MAB}
\end{figure}

To demonstrate how this can be used to train GANs, consider the usual value function~\citep{originalGAN}:
\begin{align}
    V(D, G) = \Esp_{p(x)}[\log(D(x))] + \Esp_{p(z)}[\log(1 - D(G(z)))].
    \label{eq:gan_loss}
\end{align}
On each episode $t$, given the mixture of discriminators $\Pi(t)$, each discriminator is trained by taking a gradient step to increase the expected value function
\begin{align}
    \Esp_{\Pi(t)}[V(D_i, G)] = \sum_j \pi_j(t) V(D_j, G),
    \label{eq:gan_disc_loss}
\end{align}
and the generator is trained by taking a gradient step to increase
\begin{align}
    \Esp_{\Pi(t)}[\Esp_{p(z)}[\log(D(G(z)))]].
    \label{eq:gan_gen_loss}
\end{align}
The latter corresponds to the non-saturated version of Eq.~\ref{eq:gan_disc_loss} for the generator.
The intuition is that training the generator with all the discriminators \emph{simultaneously} (as a mixture) should force the generator to fool all discriminators at the same time~\citep{GMAN}. Since each discriminator has an increasing level view of the modes distribution, they should have a complementary role. While the weaker discriminator focuses on modes coverage, the stronger discriminator ensures samples quality (showed in Section~\ref{disc_smoothness}). This should result into a better overall coverage of the modes in the input distribution.



Algorithm~\ref{alg:MAB} describes our proposed acGAN procedure.
We denote and parameterize this algorithm as $\text{acGAN}(\lambda,\alpha,\mathcal{R}_{r})$ where $\lambda \geq 0, \alpha \in (0,1)$.

\begin{algorithm}
   \caption{Generic acGAN algorithm}
   \label{alg:dist_gan}
\begin{algorithmic}[1]
 \STATE {\bfseries Given:} $N$: number of discriminators, $T_{max}$: time steps, $T_{warmup}$: warmup time, $\alpha$: moving average coefficient, $\lambda$: Boltzmann constant 
 \STATE $Q_i(0) \gets 0 ,\forall i=1,\dots,N$
 \FOR{$t=1,\dots,T_{max}$} 
 \STATE Update all discriminators $\{D_{i}\}_{i=1}^{N}$ using Eq.~\ref{eq:gan_disc_loss}
 \STATE Update the generator $G$ using Eq.~\ref{eq:gan_gen_loss}
 \IF {$t \geq T_{warmup}$} 
   \STATE Evaluate the performance of $G$ and observe a reward $\cR_i(t)$ for each discriminator $i$
   \STATE Update all values $\{Q_{i}(t)\}_{i=1}^N$ according to Eq.~\ref{eq:MA_Q_value}
   \STATE $\pi_i(t) \gets \exp^{\lambda Q_i(t)} / \sum\limits_{j=1}^N \exp^{\lambda Q_j(t)} \quad \forall i=1\dots N$
 \ENDIF
 \ENDFOR
\end{algorithmic}
\label{alg:MAB}
\end{algorithm}

\begin{remark}
    At the beginning of the training, we define a warm-up period $T_{\text{warmup}}$, prior to which we train $D_{i}$ and $G$ with a uniform probability, i.e $\pi_{i}=\frac{1}{N}, \forall i=1,\dots,N$. In other words, we consider $\lambda = 0, \forall t \leq T_{\text{warmup}}$. This guarantees that each discriminator is updated a minimum number of times (or provides feedback a minimum number of times to the generator) and prevents one $D_j$ from dominating the others (i.e, $\pi_{j} \gg \pi_{i}, \forall i \neq j$) at the beginning of the training. Without this safeguard, the remaining weights $\pi_i, i \neq j$ would hardly recover a significant probability and the generator may never get informative gradient from the corresponding discriminator. Note that warm-ups are not uncommon either in bandits algorithm, e.g. for adding robustness to the tails of reward distributions~\citep{Baransi2014}.
\end{remark}


\subsection{Reward shaping}

In order to provide meaningful feedback for learning efficient mixtures of discriminators, we consider different reward functions to generate $\cR_i(t)$. We argue that progress (i.e., the learning slope~\citep{teacher_student,AutomatedCuri}) of the generator is a more sensible way to evaluate our policy. 
Let $\theta(t)$ be the generator parameters at episode $t$.
We define the two following quantities for measuring generator progress:
\begin{align}
\label{eq:reward_quality_S}
\mathcal{R}_i^{\mathcal{S}}(t) &= \Esp_{p(z)}[D_{i}(G(z; \theta(t))) \nonumber \\ & \qquad\qquad - D_i(G(z; \theta(t - 1)))],\\
\label{eq:reward_quality_2}
\mathcal{R}_i^{\mathcal{V}}(t) &= \Esp_{p(z)}[V(D_i, G(z; \theta(t))) \nonumber \\ & \qquad\qquad - V(D_i, G(z; \theta(t - 1)))].
\end{align}
The former measures the progress of the generator with respect to the discriminator $i$ score $D_{i}(\cdot)$, while the latter assess the change in the loss function (Eq.~\ref{eq:gan_loss}). Since the change in the quality sample (Eq~.\ref{eq:reward_quality_S}) led to better performance than the change in the loss function (Eq~.\ref{eq:reward_quality_2}), all our experiments (see Section.~\ref{sec:experiments}) use Eq~.\ref{eq:reward_quality_S}.

\subsection{Connection to existing methods}





Interestingly, some existing methods in the GAN literature can be seen as a specific case of acGAN:

\paragraph{GMAN:}
The original GMAN~\citep{GMAN} algorithm can be recovered by setting $\alpha=1$ and taking the loss function to be the reward $\mathcal{R}_{i}(t)=V(D_{i},G)$.
Note how the authors of GMAN call their algorithm GMAN-$\lambda$, where $\lambda$ is also the Boltzmann coefficient.

\paragraph{Uniform:}
The uniform case is defined by assigning a fixed uniform probability for each discriminator $D_i$: 
\begin{align*}
    \pi_{i}(t)=\frac{1}{N}, \quad \forall t\in \Nat.
\end{align*}
This corresponds to Eq.~\ref{eq:boltzman_policy} with $\lambda=0$.


To support the results of our theoretical work, we conducted a set of experiments which we describe below.

\section{Experiments}
\label{sec:experiments}
In this section, we first give an understanding of how each discriminator provides informative feedback to the generator. We then compare our proposed approach (acGAN) against existing methods from the literature.

\subsection{Retaining mode information through weaker capacity discriminators and smoothness}
\label{disc_smoothness}

We begin by analyzing the gradient norm of the discriminator networks and we show that weak capacity discriminators are \emph{smoother} than strong discriminators. This property corresponds to a "coarse-grained" representation of the distribution, which allows the generator to recover missing modes. 
We further show we can increase the smoothness of a weak discriminator by corrupting its inputs with white noise. This results in an increase of the discriminator's entropy (see Supplementary Material for more details) and hence smoother gradient signal. 

\subsubsection{Weak Discriminators: a way to retain modes}

We now highlight the role of weaker capacity discriminators. To this extent, we performed the following experiments on the 8 Gaussian synthetic dataset:
\begin{itemize}[leftmargin=*]
    \item We pretrained the generator (with $3$ dense layers of $400$ units with ReLU activation layers except for the last layer) with one discriminator on only $2$ of the original $8$ modes.  
    \item We trained a (vanilla) GAN on all $8$ Gaussian components, initializing with the $2$-mode generator above. The discriminator had $3$ dense layers of $400$ units (ReLU hidden activation layers).
    \item We trained acGAN with the generator initialized with the $2$-mode generator (as with vanilla GAN). We considered $3$ discriminators, with $1$, $2$ and $3$ dense layers respectively (same activation scheme as previously applies here).
\end{itemize}

\begin{figure}[H]
    \hspace*{-0.5cm} 
    \centering
    \includegraphics[width=9cm,height=3cm]{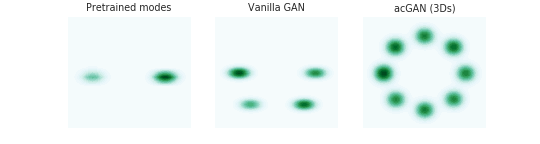}
    \caption{Modes used for pretraining the generator (left) and modes recovered by Vanilla GAN (middle) and acGAN (right). The more modes the better.
    }
    \label{fig:mode_recovering}
\end{figure}

\begin{figure}[H]
    \hspace*{-0.1cm}  
    \centering
    \includegraphics[width=8.5cm,height=2.25cm]{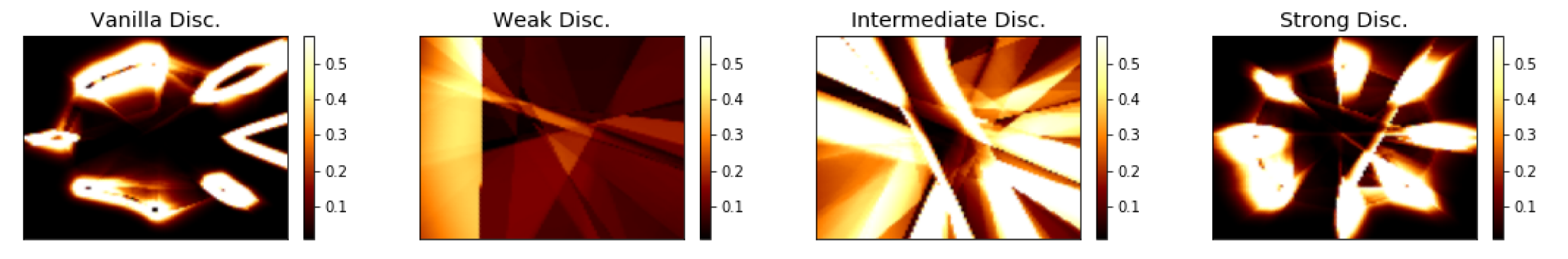}
    \caption{Gradient norm of each discriminator with respect to the input. We clipped the magnitude with respect to the weaker discriminator range. Since weaker discriminators are smoother by construction, they help the generator to recover missing modes. On the other hand, vanilla GAN can hardly recover modes due to its vanishing gradient.}
    \label{fig:recovering_modes_grad_norm}
\end{figure}

Results (Fig.~\ref{fig:mode_recovering}) show the Vanilla GAN could only retrieve $2$ additional modes, while acGAN recovered all ($8$) modes.
We examined the gradients provided by the discriminators using a density plot (Fig.~\ref{fig:recovering_modes_grad_norm}) of the gradient norm for each discriminator with respect to the input, i.e., $||\nabla_{X} D(X) ||_2$ for $X \in [-2,2]^2$. Observe that there is a clear progression from a stronger discriminator with more distinct, higher gradients to the weaker discriminator smoother gradients. Additionally, note that the discriminator from the vanilla GAN, which has very high gradient norm values, has gradients for modes not present in the generator: the discriminator has information useful for learning about these missing modes, but the generator does not learn these modes due to vanishing gradients.
Our results support both our original hypothesis that missing modes are due to vanishing gradients and that using a coarse-grain discriminator can be used to recover missing modes.
To provide further insight, we show the evolution of the gradient norm of each discriminator at training time in the Supplementary Material.
We also note that the discontinuities in the gradients is due to the ReLU activation partitioning the subspace through overlapping half-planes, which contrasts the smooth decay of hyperbolic tangent and sigmoid\footnote{$\sigma(y) = 1 / 1 + e^{-y}$} nonlinearities, and we further explore the effect of different nonlinear activation layers on the gradient norm of the weak discriminator in the Supplementary Material.


\subsection{Performance of acGAN against existing baselines}

In this section, we evaluate the performance of our proposed method (acGAN), on various datasets. All experiments consider the reward shown in Eq.~\ref{eq:reward_quality_S}. We first conducted a sanity check on 2 mode-dropping datasets: synthetic data consisting of a mixture of 25 Gaussians and Stacked-MNIST with 1000 modes. We then tested it on CIFAR10 and finally show generated samples on celebA dataset (see Supplementary Material). 
We aim to analyze specific properties such as diversity of generated samples and quality in terms of ~\citep[FID,][]{FID} score when available, along with convergence of the method (how fast it reaches its minimum FID score). Additionally, our results hint at the emergence of a curriculum during the training process.
\\
All parameters used to obtain the results can be found in the Supplementary Material.
We split the batch of inputs between discriminators. We abuse of language with the term \textit{epoch}, which in the context of the current paper means that the generator has been trained on a number of iterations equivalent to an epoch. For example, CIFAR-10 has 50,000 training images and, assuming a batch size of 64, one epoch represents roughly 781 iterations for the generator.



\begin{table}
\hspace*{-0.8cm}
\centering
\resizebox{0.45\textwidth}{!}{
\begin{tabular}{l|ccc}
 \hline
                   &   FD  & Modes & Quality samples \\  \hline
\multicolumn{1}{c|}{ \text{Vanilla GAN}}  &  $7.28$  & $17$     & $88\%$   \\ 
\multicolumn{1}{c|}{ \text{Uniform (3D)}}      &  $6.64$     & $20$     &  $93.4\%$ \\
\multicolumn{1}{c|}{ \text{ acGAN (3D) }}     &  $6.65$       &  $25$ & $92.9\%$       \\ \hline
\end{tabular}}
\caption{Results on the Gaussian mixture synthetic data. Our method acGAN could cover allc 25 modes.}
\label{tab:toy} 
\end{table}

\begin{figure}
    \centering
    \includegraphics[scale=0.65, clip, trim={60 20 10 0}]{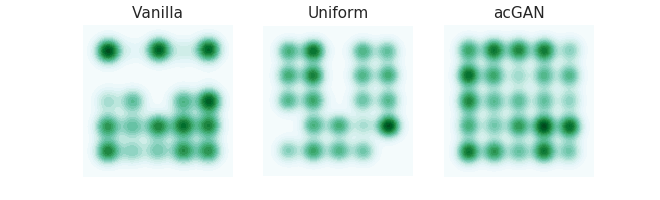}
    \caption{KDE plots of the modes recovered by each examined approach with 3 discriminators.}
    \label{fig:KDE}
\end{figure}

\subsubsection{Synthetic Gaussian mixture dataset}
The synthetic dataset is composed of 25 bivariate Gaussian mixtures arranged in a two-dimensional grid.
We launched a single run of 15 epochs for all methods with 3 discriminators. We report 3 measures in Table~\ref{tab:toy}: the Fr\'echet Distance (FD), the number of recovered modes and the proportion of high quality samples (which is the proportion of samples covering a mode). More details on those metrics can be found in the Supplementary Material. 

We compared the performance of our proposed methods to that of the Uniform algorithm and of the vanilla GAN~\citep{originalGAN}. Our proposed methods could cover the 25 modes. KDE plots for the 3 discriminators case are shown in Fig. \ref{fig:KDE}.

\begin{table}
\centering
\resizebox{0.45\textwidth}{!}{
\begin{tabular}{l|cc}
    \hline
                    &   Modes (max 1000)  & KL   \\  \cline{1-3}
\multicolumn{1}{c|}{ DCGAN~\citep{DCGAN} }  &  $99.0$  &  $3.40$        \\
\multicolumn{1}{c|}{ ALI~\citep{ALI} }      &  $16.0$    & $5.40$   \\
\multicolumn{1}{c|}{ Unrolled GAN \citep{UnrolledGAN} }          &  $48.7$    &  $4.32$     \\ 
\multicolumn{1}{c|}{ VEEGAN~\citep{VEEGAN} }     &  $150.0$      &  $2.95$ \\ 
\multicolumn{1}{c|}{ PacGAN~\citep{PacGAN}}  &  $1000.0 \pm 0.00$ & $0.06 \pm 1.0e^{-2} $\\
\multicolumn{1}{c|}{ GAN+MINE~\citep{MINE}}  &  $1000.0 \pm 0.00$ & $0.05 \pm 6.3e^{-3} $\\\hline
\multicolumn{1}{c|}{ acGAN (3D) }     &    $1000.0 \pm 0.00$      &  $7.4e^{-2} \pm 0.0$        \\
\multicolumn{1}{c|}{ acGAN (5D) }     &   $1000.0 \pm 0.00$       &  $9.65e^{-2} \pm 0.0$        \\
\hline
\end{tabular}
}
\caption{Number of modes covered and Kullback-Leiber divergence between the real and generated distributions on Stacked-MNIST. acGAN could recover  the 1000 modes.}
\label{tab:stacked_mnist}
\end{table}

\begin{figure}
\begin{minipage}{0.6\textwidth}
\begin{minipage}[b]{0.45\textwidth}
\hspace{-0.6cm}  
\centering
\captionsetup{type=figure}
\includegraphics[width=4.5cm,height=4.5cm]{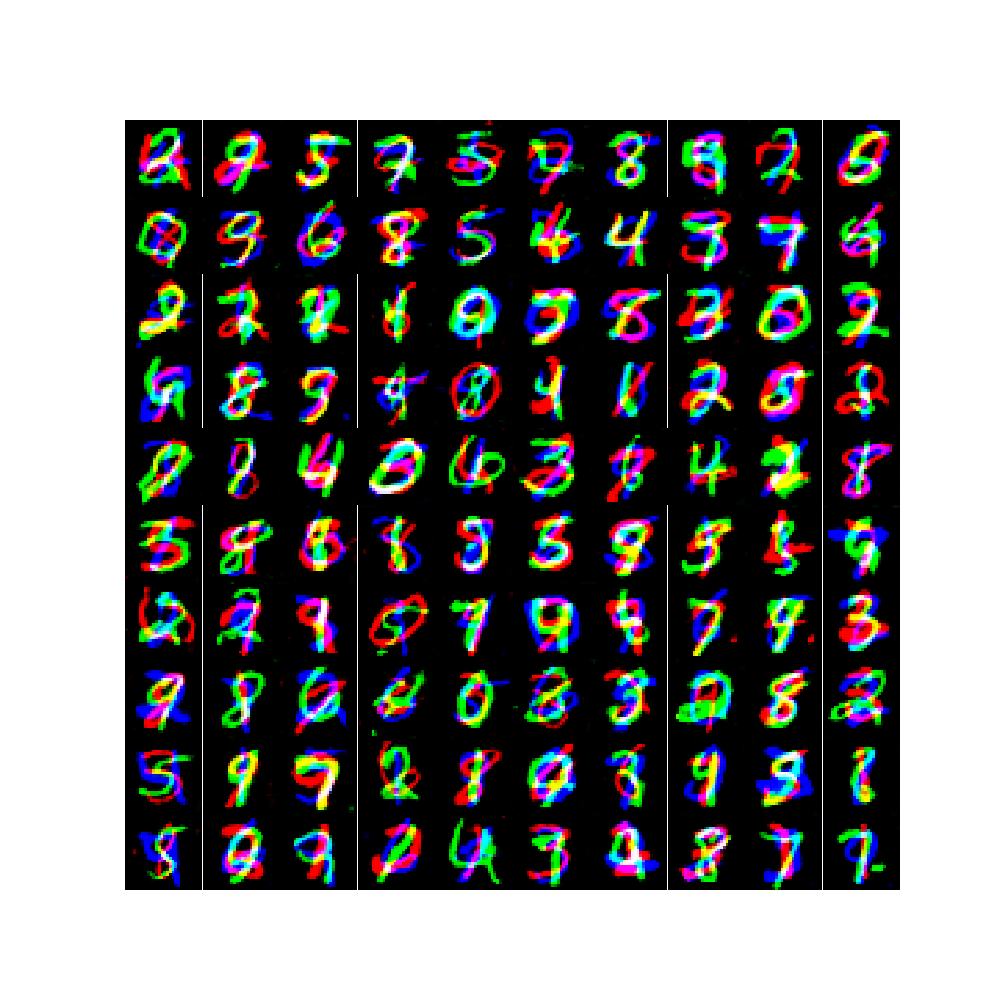}
\end{minipage}
\hspace{-1.0cm}
\begin{minipage}[b]{0.45\textwidth}
\centering
\captionsetup{type=figure}
\includegraphics[width=4.5cm,height=4.5cm]{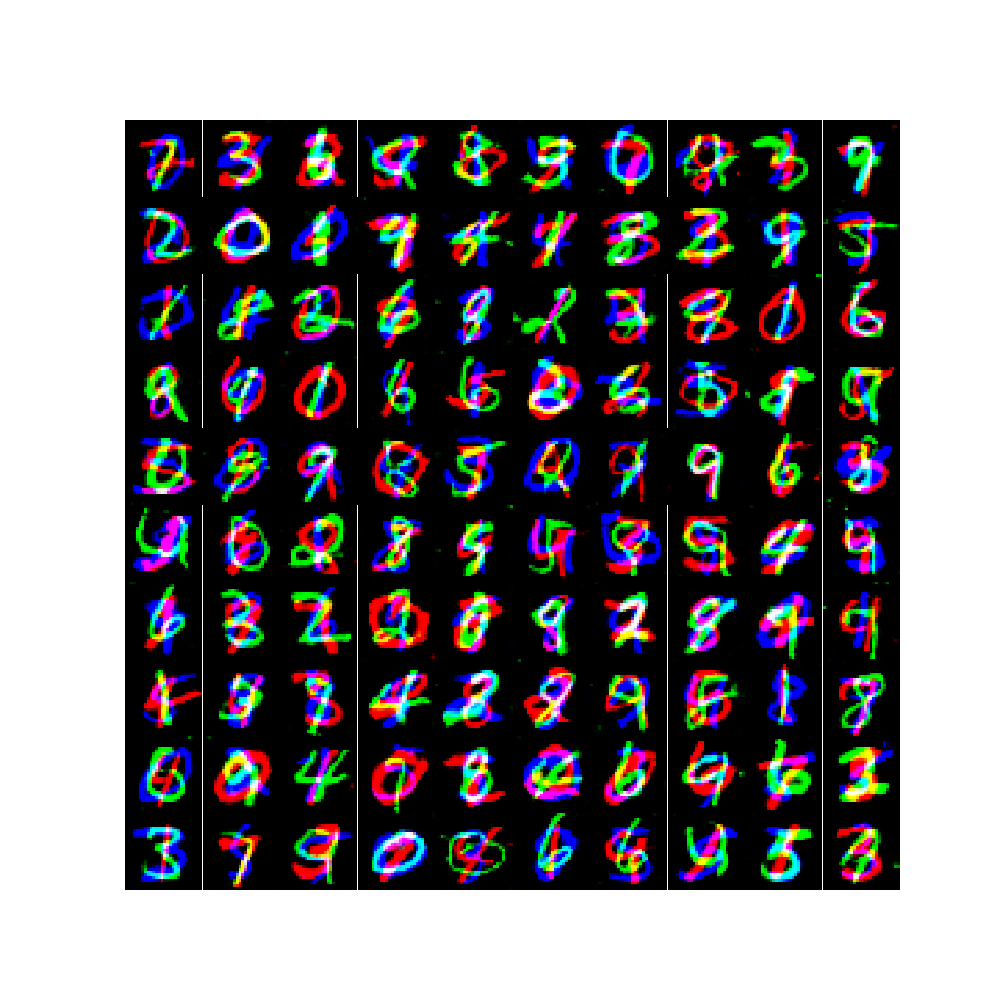}
\end{minipage}
\end{minipage}
\captionof{figure}{Stacked-MNIST generated samples for acGAN with 3 (left) and 5 (right) discriminators.}
\label{fig:stacked_mnist_samples}
\end{figure}

\subsubsection{Stacked-MNIST}

We use the Stacked-MNIST dataset~\citep{srivastava2017veegan} to measure the mode coverage of our proposed approach. The dataset is generated by stacking 3 randomly selected digits from the MNIST dataset: one on each RGB channel to produce a final $28 \times 28 \times 3$ RGB tensor. The dataset has 128,000 training images and is assumed to have $10^3$ modes. Results of our experiments are shown in Table~\ref{tab:stacked_mnist}. 

We report our results (averaged over 10 runs) in Table~\ref{tab:stacked_mnist} and compare them with other existing baselines in the literature. Our method could recover all 1000 modes like PaCGAN~\citep{PacGAN} and MINE~\citep{MINE}; these two approaches either increase the dimensionality of the generator inputs either by packing multiple samples or by adding a latent code vector which helps overcoming mode collapse. Generated samples are shown in Fig.~\ref{fig:stacked_mnist_samples}, our results further verify our hypothesis that acGAN is a sensible approach to ensuring good mode coverage and sample quality.



\begin{figure}[H]
  \hspace*{-0.85cm}  
    \centering
   \includegraphics[width=10cm,height=5cm]{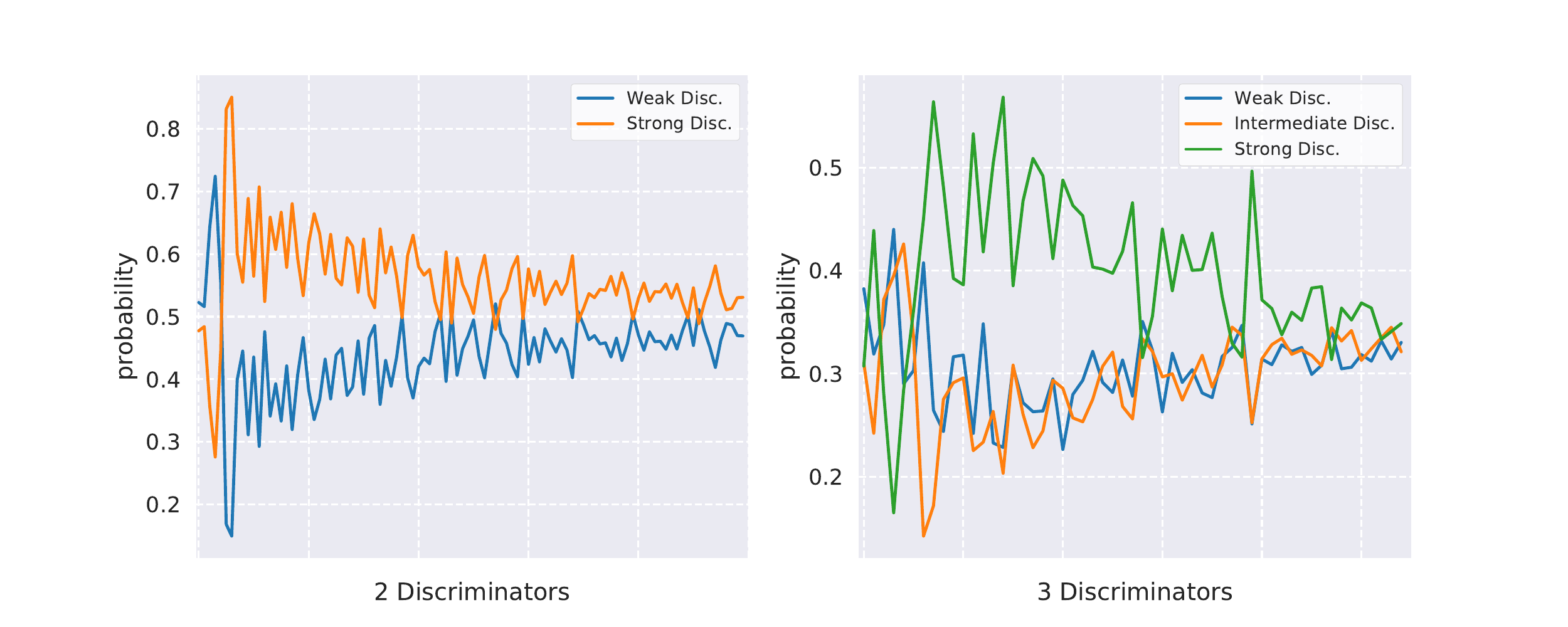}
    \caption{Weight $\pi_{i}$ of each discriminator over the training epochs. We can see phase switching at the beginning where each discriminator's weight is dominating before eventually converging to a uniform distribution.}
    \label{fig:sampling_proba_cifar10}
\end{figure}

\subsubsection{CIFAR-10}

We conducted an in-depth study of acGAN's performance on CIFAR-10 by running experiments on 5 independent seeds for 50 epochs each.

We found a particular pattern in the acGAN's learning process: it consists of distinct regimes where one discriminator's weight $\pi_{i}$ dominates over the others. To illustrate this, we averaged the sampling probability of each discriminator over every 200 iterations and plotted results in Fig. \ref{fig:sampling_proba_cifar10} for $2$ and $3$ discriminators, respectively. The reported curves suggest that, for $N=2$ discriminators, the weakest discriminator network is often sampled at the beginning until the generator $G$ learns enough from it, at which point it begins to use the stronger discriminator more often. Note how the strong discriminator is sampled more frequently than the weak one. In fact, because the generator needs to produce samples of higher quality to fool the strong discriminator, training with the latter might take longer as opposed to using weaker discriminators (which are more lenient). By the end of training, all discriminators are being used in equal proportions, meaning that every discriminator plays a complementary role from mode coverage to quality samples. A similar pattern is observed for the $3$-discriminators case. 

To assess the quality of produced results, we report the minimum Fr\'echet Inception Distance~\citep[FID,][]{FID} (and corresponding epoch) reached in Table~\ref{fig:results_cifar10}. The squared FID was computed every epoch with 1,000 held-out samples at training time. As in \cite{fedus2017many}, a ResNet pre-trained on CIFAR-10 was employed to obtain representations for FID computation rather than Inception V3. Proceeding this way yields a more informative score, given that our classifier was trained on the same data as the generative models. Details on the FID score can be found in the Supplementary Material.
We compared our results to \citet{GMAN}. Since the authors reported that GMAN-1 ($\lambda=1$) had an overall better performance, we used this version in our experiments and refer to it as GMAN. 
Previously, we observed that the feedback provided to the generator is shared between all the discriminators. Especially, not all gradient comes from the strong discriminator (unlike for the Vanilla GAN). One might be concerned by a degradation of the quality samples. We show that having more discriminators leads to better mode coverage and samples quality (see the FID curves for an increasing number of discriminators in the Supplementary Material).
Overall, we noticed that acGAN achieved the best FID score when compared to the baseline as presented in Fig.~\ref{fig:comparison_3_disc} and \ref{fig:comparison_5_disc}  (plots are shown in a larger format in in the Supplementary Material).
GMAN performed worse than expected and increasing the number of discriminators did not significantly improve its FID score. We suspect that the original loss function of the GAN (which is equivalent to the Jensen-Shannon divergence minimization) is not a good signal to assess the progress of $G$. Indeed, \cite{WGAN} argued and introduced a toy example showing that this version of adversarial nets is not informative when there is little overlap between the supports of the true and approximate distributions, as commonly seen at the beginning of the training process. Finally, not keeping a moving average via a $Q$-value can lead to high variance.




\begin{figure}[h]
    \hspace*{-0.5cm}  
    \centering
    \includegraphics[width=9cm,height=6cm]{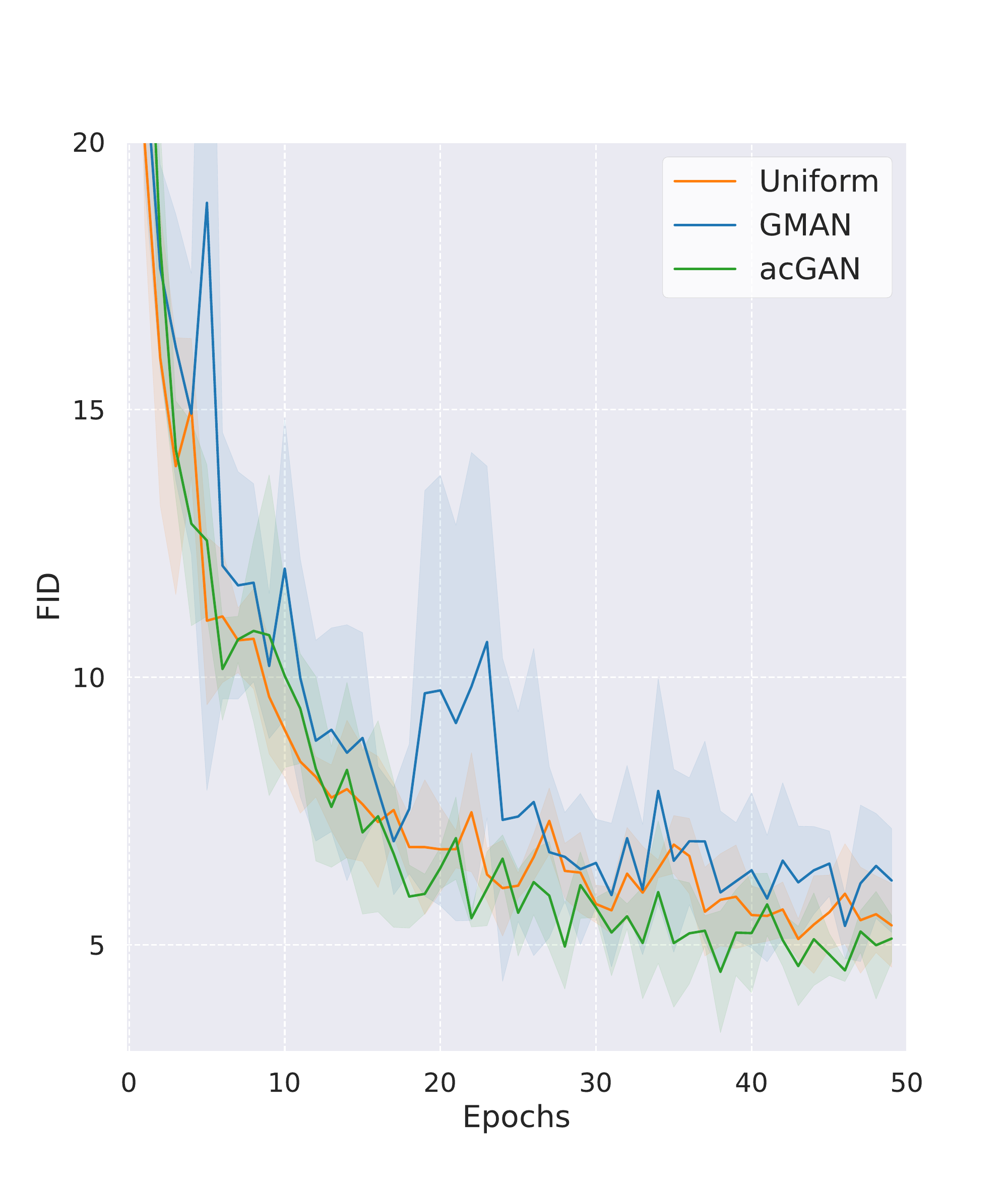}
    \caption{FID scores computed with 1,000 samples at the end of each epoch for different methods with 3 discriminators. acGAN outperforms the baselines Uniform and GMAN.}
    \label{fig:comparison_3_disc}
\end{figure}

\begin{figure}[h]
    \hspace*{-0.5cm}  
    \centering
    \includegraphics[width=9cm,height=6cm]{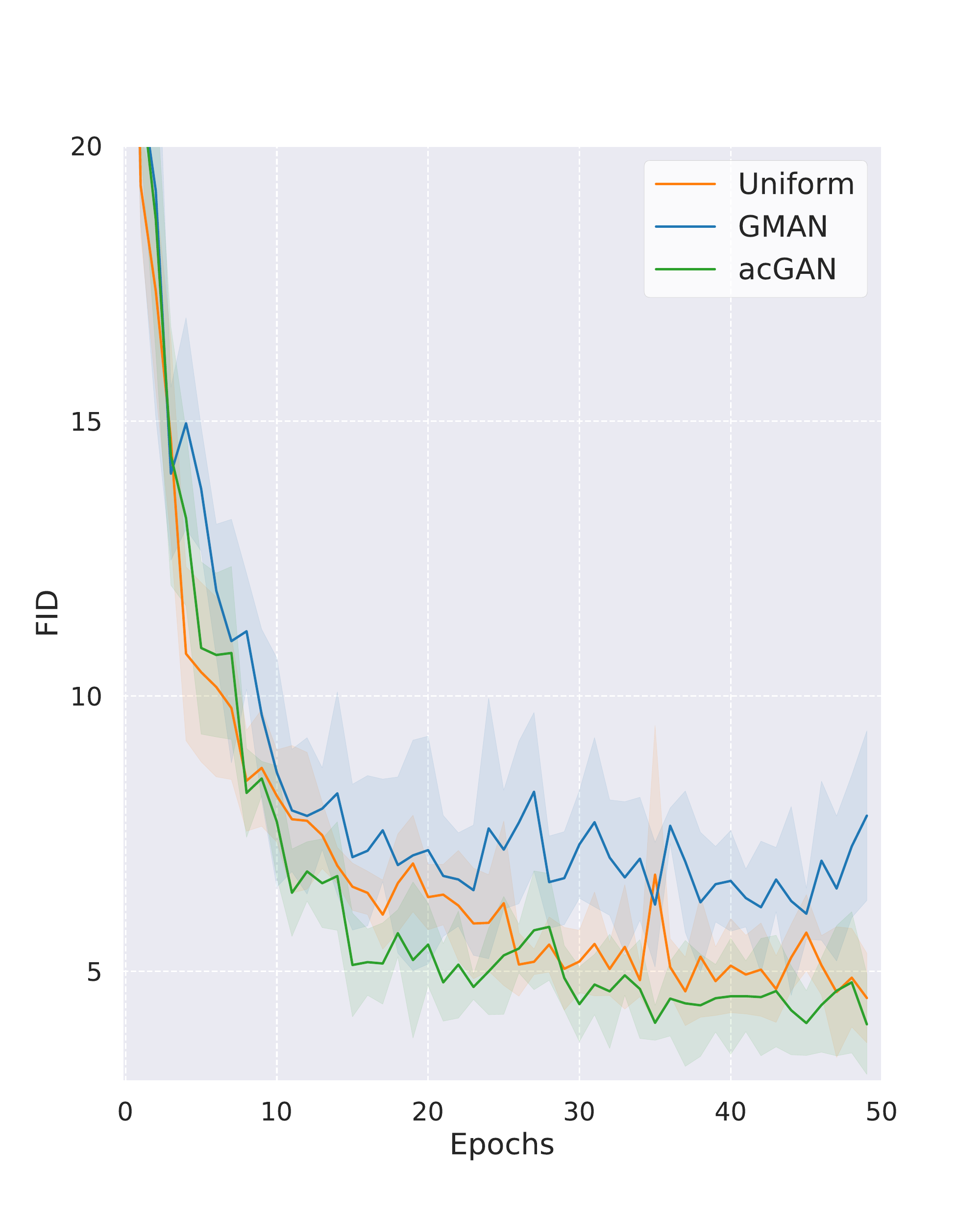}
    \caption{ FID curves with 5 discriminators. acGAN presented earlier convergence and reached lower FID values.}
    \label{fig:comparison_5_disc}
\end{figure}

\begin{table}[h!]
\centering
\resizebox{0.45\textwidth}{!}{
\begin{tabular}{l|c|cc|c}
\hline
   &    & \multicolumn{2}{c|}{Best FID \textit{(epoch)}  }  & Mean Best FID     \\ \hline    
   & Vanilla GAN &  \multicolumn{2}{c|}{\emph{5.02} \textit{(20)} - \emph{5.28} \textit{(27)} -  \textbf{\emph{4.27} \textit{(30)}} - \emph{4.80} \textit{(34)} - \emph{4.63} \textit{(41)} }   & \emph{4.80 }   \\  \hline  
& WGAN-GP\footnotemark &  \multicolumn{2}{c|}{\emph{4.29} \textit{(43)} - \emph{4.24} \textit{(28)} -  \emph{3.98} \textit{(47)} - \emph{3.99} \textit{(37)} - \textbf{\emph{3.93} \textit{(50)}} }   & \emph{4.08 }   \\  \hline  
\multirow{3}*{\rotatebox[origin=c]{90}{3 Disc}} &   Uniform    &       \multicolumn{2}{c|}{\emph{4.18} \textit{(20)} -  \textbf{\emph{4.07} \textit{(39)}} -  \emph{4.35} \textit{(45)} - \emph{5.07} \textit{(30)} - \emph{4.39} \textit{(47)} }   	& \emph{4.41 }	\\ 
  &   GMAN    		  &       \multicolumn{2}{c|}{ \textbf{\emph{3.87} \textit{(43)}} - \emph{4.05} \textit{(46)} -  \emph{5.24} \textit{(42)} - \emph{5.71} \textit{(42)} - \emph{4.10} \textit{(22)} }   	& \emph{4.59 }	\\ 
  
  
   &   acGAN    		  &  \multicolumn{2}{c|}{\emph{3.93} \textit{(39)} - \emph{3.57} \textit{(38)} -  \emph{4.25} \textit{(42)} - \emph{3.43} \textit{(40)} - \textbf{\emph{3.11} \textit{(43)}} }   	& \textbf{3.66}	\\  \hline
   
\multirow{3}*{\rotatebox[origin=c]{90}{5 Disc}} &   Uniform   &  \multicolumn{2}{c|}{ \textbf{\emph{3.\textbf{42} \textit{(47)}}} - \emph{3.69} \textit{(49)} -  \emph{4.37} \textit{(37)} - \emph{3.64} \textit{(37)} - \emph{3.47} \textit{(40)} }   	& \emph{3.72}	\\ 

&   GMAN    		  &       \multicolumn{2}{c|}{\emph{4.58} \textit{(44)} - \emph{4.40} \textit{(20)} -  \textbf{\emph{3.91} \textit{(47)}} - \emph{4.81} \textit{(25)} - \emph{4.42} \textit{(38)} }   	& \emph{4.42}	\\

  
 &   acGAN    		  &       \multicolumn{2}{c|}{\emph{3.62} \textit{(35)} -  \textbf{\emph{2.62} \textit{(49)}}  -  \emph{4.14} \textit{(35)} - \emph{2.66} \textit{(42)} - \emph{3.67} \textit{(34)} }   	& \textbf{3.34}	\\ 
\hline
\end{tabular}}
\caption{Best FID scores on CIFAR-10 computed on 1,000 samples during training time (lower is better).}
\label{fig:results_cifar10}
\end{table}

\footnotetext{We replaced the batch norm layer with instance norm}


\section{Conclusion}
\label{sec:conclusion}

In this work, we model the training of the generator against discriminators of increasing complexity within a one-student/multiple-teachers paradigm. We address this mixture-of-experts problem under the adversarial bandit setting with full-information, where we rely on the Hedge algorithm to learn the weights assigned to each discriminator in the mixture. Since designing a suitable reward function is a key ingredient to control the shape of the learned policy, we examined two sensible reward functions which relied on sample quality and the GAN loss function. We empirically found the high quality sample reward (Eq. \ref{eq:reward_quality_S}) to yield the best results. Keeping a moving average on the rewards helped smoothing the weights put on discriminators and resulted in a more stable mixture. 

Then, we demonstrated a complementary regulation mechanism between weak and strong discriminators. While weaker discriminators enjoy smoother properties and provide more informative feedback to the generator, stronger discriminators focus one finer grain detail to ensure sample quality.

Finally, we conducted a series of experiments to show the emergence of a curriculum during the training process. That is, lower-capacity discriminators have higher weights at the beginning but, as the training progresses, higher weights are allocated to higher-capacity discriminators. We showed how existing algorithms could be recovered from our model via the $Q$-value. The performed experiments showed that our proposed approach leads to an earlier convergence and a better FID score compared to existing baselines in the field, i.e. Uniform and GMAN. 

As a direction for future investigation, approaches not relying on the adversarial framework could be investigated to model the non-stationarity of the reward distributions. For example, finding a meaningful representation for the state of the generator could allow the use of contextual bandits algorithms.

\bibliography{bibliography}
\bibliographystyle{aaai}

\newpage
\onecolumn

\section*{Supplementary Material}
\begin{figure}[h!]
    \centering
    \includegraphics[width=\linewidth, height=0.5\linewidth, clip, trim={95 60 85 0}]{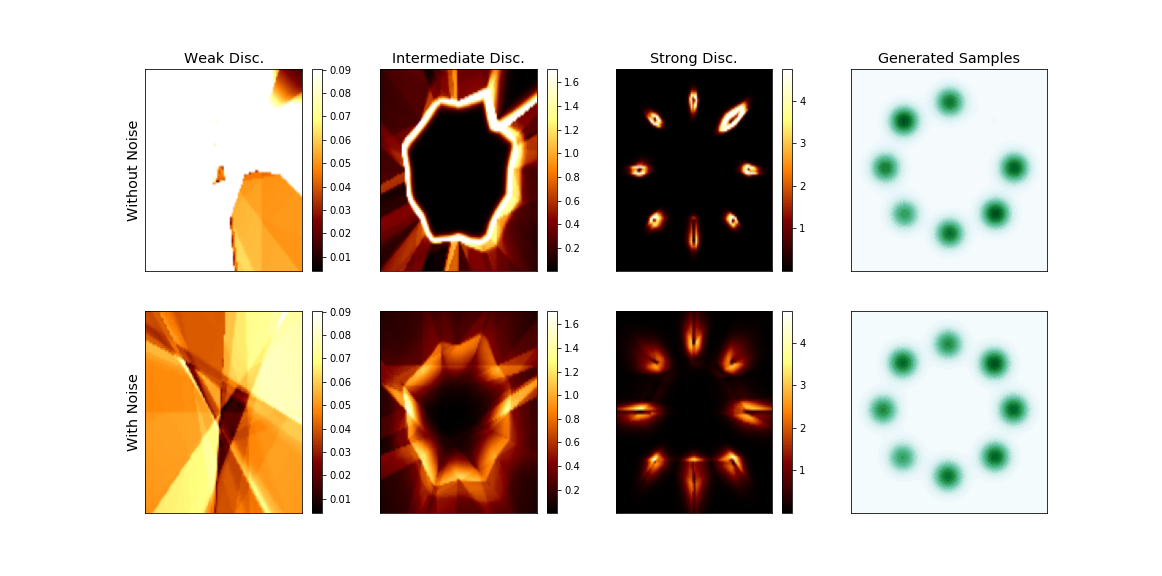}
    \caption{Adding noise (bottom row) reduces gradient norm magnitude of each discriminator. This increases their smoothness properties and helps recovering modes of the distribution. We clipped the gradient magnitude with respect to the corresponding discriminator corrupted with noise.}
    \label{fig:noise_effect}
\end{figure}

\begin{figure}[H]
    \centering
    \includegraphics[width=\linewidth, height=0.5\linewidth, clip, trim={95 60 85 0}]{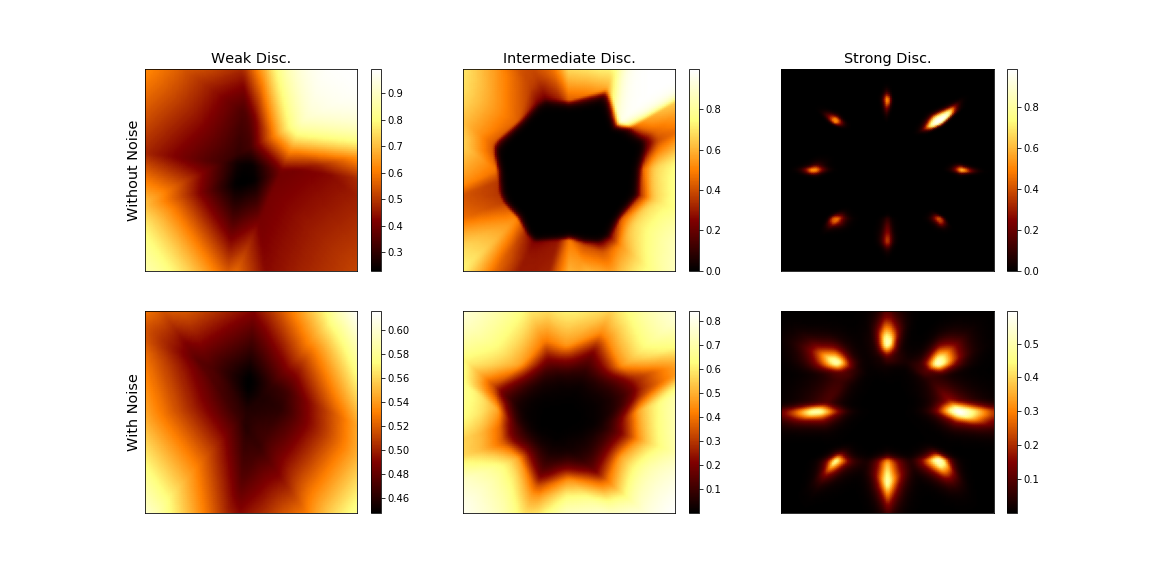}
    \caption{Probability $D(x)$ for each discriminator without (top) and with (bottom) white Gaussian noise. Noise tends to smooth their decision boundary and increase their entropy. That helps to provide more informative gradient to the generator.}
    \label{fig:prob_all_D}
\end{figure}

\subsubsection{Regularizing the discriminator through additive white noise} 

\begin{figure}[H]
    \centering
    \includegraphics[width=12cm,height=3.5cm]{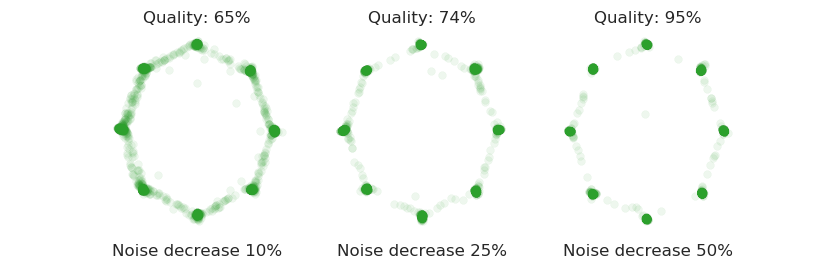}
    \caption{As we exponentially decay the noise, samples quality increase (2500 samples are plotted).}
    \label{fig:noise_evolution}
\end{figure}

As was explored in \citet{arjovsky2017towards}, one way to stabilize GAN training is corrupting the input of the network with additive white Gaussian noise of the form $N(0,\sigma)$. 
Here, we explore smoothing the discriminator by using the noise, We ran the following experiment in order to illustrate the mechanism. We (once again) train $G$ on the 8 Gaussian synthetic dataset with 3 discriminators (1,2 and 3 hidden layers of $256$ units) both with and without adding independent Gaussian noise to the discriminator's input. A noticeable downside of feeding corrupted inputs to the network is the degradation of samples' quality: the so-called salt and pepper effect becomes more visible as the discriminators train. To solve this issue, we decay the noise at time step $t$ by a multiplicative coefficient: $\exp{\frac{t}{C}}$, where $C>0$ is a real constant controlling the noise reduction speed. Initial Gaussian noise was picked to be of the form $N(0,\sigma_i)$, with variances of $\sigma_1=0.06$, $\sigma_2=0.04$, $\sigma_3=0.02$, 
for $i=1$ being the weakest discriminator and $i=3$ the strongest. Adding white noise increases the entropy (read uncertainty) of the discriminator (a proof is shown in the Supplementary Material) and tends to smooth its decision boundary (see the probability and gradient norm values in Figs.~\ref{fig:prob_all_D} and \ref{fig:noise_effect}). 
Fitting a discriminator to uncorrupted input is prone to faster overfitting as opposed to training on noisy data when fixing the number of parameters, a great illustration of which is provided by \cite{gu2008smoothing}. Empirical results are shown in Fig. \ref{fig:noise_effect}. We see that by corrupting the real data we manage to cover all 8 modes and the sample quality is conserved by decaying the variance of the noise. The evolution of generated data points is shown in Fig.~\ref{fig:noise_evolution}.
 

\subsection{Effect of different nonlinear activation layer on the weak discriminator's smoothness} 

\begin{figure}[H]
    \centering
    \includegraphics[width=15cm,height=4cm]{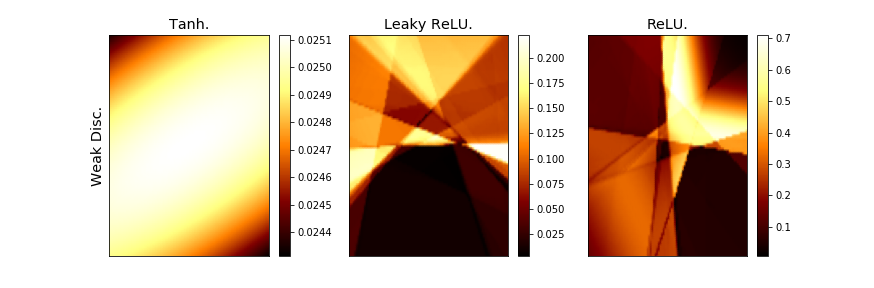}
    \caption{Although Tanh (left) presents smoother partition of the subspace than LeakyReLU (middle) and ReLU (right), it seems to have weak gradient signal (small gradient norm magnitude).}
    \label{fig:nonlinear_activation}
\end{figure}

In this section, we aim to illustrate the effect of 3 nonlinear activation layers (Tanh, Leaky RelU and ReLU) on the gradient norm of the discriminator. We ran the training of the generator with 3 discriminator (using Soft-acGAN) on the 8 Gaussian dataset. For some performance issue, we just replace the activation layer of the weak discriminator with respectively the 3 above mentionned activation layers and let the other discriminator with ReLU. Fig.~\ref{fig:nonlinear_activation} shows the gradient norm of the weak discriminator on the whole space $[-2,2]^2$. We see that Tanh has a very uniform gradient norm across the space while ReLU is the most discontinuous. Leaky ReLU has an intermediate pattern. Yet, Tanh seems to have flat behaviour (very small magnitude), this may be due to the Tanh function that has very low gradient signal at the extremity (indeed, we witness very poor performance with that activation layer). Leaky ReLU is less discontinuous although it also partitions the subspace in the same way as ReLU.

\subsection{Evolution of the gradient norm during the training} \label{evolution_grad_norm}
\begin{figure}[H]
    \centering
    \includegraphics[width=1.0\linewidth]{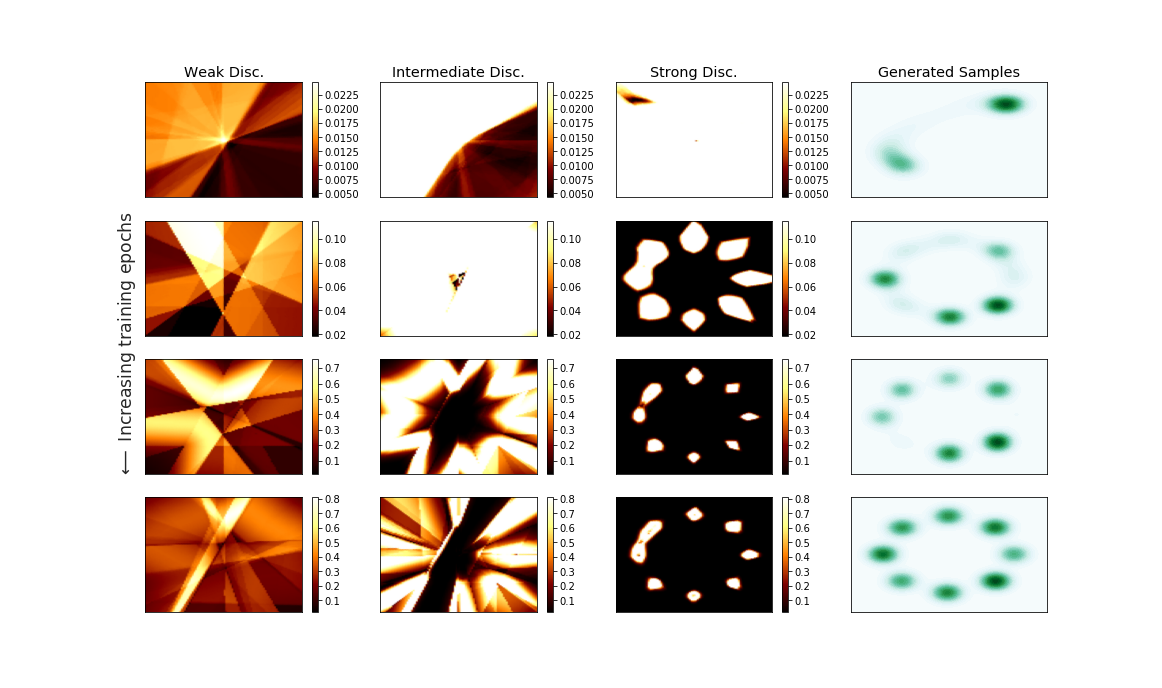}
    \caption{Evolution of the gradient norm for each discriminator and samples generated (last column). The generator recovers modes thanks to the gradients provided by the weak and intermediate discriminators. Each discriminator in turn evolves to learn its coarse to fine-grained representation of the data. Note also that the strong discriminator has a good representation of all the modes before the generator has learned them, indicating that mode dropping in this setting is not due to those modes being absent in the discriminator. We have clipped the gradient range with respect to the weak discriminator of the corresponding row.}
    \label{fig:evolution_gradient_norm}
\end{figure}

In this section, we show the evolution of the gradient norm of each discriminator throughout the training process (results shown in Fig.~\ref{fig:evolution_gradient_norm}). We see at the beginning of the training (first row) as the generator has just learned the top left modes, discriminator has flat behavior on the bottom right part of the subspace and has higher gradient norm on the top left part. A the training process, we see that missing modes has high gradient norm (second row third column). Finally, at the end when the generator has learned all the modes the weak discriminator seems to have more uniform gradient norm on the space while strong discriminator has equal gradient norm value at each modes locations.

\subsection{Regularizing the discriminator through additive noise}




\paragraph{Increasing entropy with additive Gaussian noise.} 


In order to have discriminator networks with varying degrees of strength, we first resorted to nested architectures: for instance, the stronger discriminator should have a more complex architecture than the weaker. Moreover, we proceeded to corrupt the inputs with additive Gaussian white noise. Formally, to the input matrix $X_i \in \mathcal{X}$ of the discriminator $D_i$ we added $\varepsilon_i \sim \mathcal{N}(0,\sigma_i^2)$, thus creating new input $Y_i=X_i+\varepsilon_i$ which was then fed to the discriminator. For practical purposes, noise for image data should be on a bounded support $R(\varepsilon_i) \subseteq R(X_i)$ in order to obtain meaningful RGB values.\\
Letting the weaker discriminators train on inputs corrupted with a Gaussian noise with larger variance allows the network to learn a high-level representation of the dataset, while feeding uncorrupted inputs will let the corresponding $D_i$ to specialize. This tradeoff between sample space coverage and estimation accuracy is known as the spike-and-slab prior and is frequently used in Bayesian variable selection methods similar to the one proposed by \citep{mitchell1988bayesian}.\\
Consider the following relation, known in information theory as the entropy power inequality (EPI). Let $X$ be a continuous, real-valued and independent random vector on a bounded support and $\varepsilon\sim \mathcal{N}(0,\sigma^2)$, both of dimension $n$:
\begin{equation}
    e^{\frac{2}{n}h(X+\varepsilon)}\geq e^{\frac{2}{n}h(X)}+e^{\frac{2}{n}h(\varepsilon)}\;.
    \label{eq:EPI}
\end{equation}
Applying logarithm on both sides and using $\log(a+b)=\log(a)+\log(1+\frac{b}{a})$, we get an expression for the entropy of the sum $X$ and $\varepsilon$:
\begin{equation}
\begin{split}
     \frac{2}{n}h(X+\varepsilon)&\geq \log(e^{\frac{2}{n}h(X)}+e^{\frac{2}{n}h(\varepsilon)})\\
    &= \frac{2}{n}h(X)+\log(1+e^{\frac{2}{n}(h(\varepsilon)-h(X)})\\
    &\geq \frac{2}{n}h(X)+\log(1+e^{\frac{2}{n}(\log{(\sigma \sqrt{2 \pi e})}-\log(b-a)})\\
    &\geq \frac{2}{n}h(X)+\log\bigg(1+\bigg[\frac{\sigma \sqrt{2 \pi e}}{(b-a)}\bigg]^\frac{2}{n}\bigg)\\
    &\geq \frac{2}{n}h(X),
    \label{eq:entropy_x_noise}
\end{split}
\end{equation}
from which it follows that adding i.i.d. Gaussian noise to the inputs increases the total entropy of the data. Here we used the fact that $\log(1+\exp(x))\geq 0 $ for all $x \in \mathbb{R}$ and that the uniform distribution $U$ has the maximal entropy over all continuous random variables with bounded support $R(U)=(a,b)$. The quantity $\log\frac{\sigma \sqrt{2 \pi e}}{(b-a)}$ controls the tightness of the bound. Because Eq.~\ref{eq:entropy_x_noise} is valid for all $\sigma$, it is necessary valid for $\sup_{\sigma >0}\log\frac{\sigma \sqrt{2 \pi e}}{(b-a)}$. Maximizing the expression shows that picking $\sigma>>0$ increases the overall entropy $h(X+\varepsilon)$ and approaches the uniform distribution.\\
Finally, recall that the entropy $h(X)=-D_{KL}(X||U)+c$ where $U\sim \text{Uniform}(a,b)$. That is, maximizing the entropy of a distribution is equivalent up to an additive constant to minimizing the Kullback-Leibler divergence between the distribution and a uniform random variable with identical support (provided adequate restrictions on the support).\\
Fitting a weak discriminator to the corrupted data should increase its capacity to generalize more than that of the stronger discriminator by acting as a regularization technique and preventing the network from overfitting.\\
Analogous mechanisms are widely used in conjunction with other learning algorithms, such as support vector machines, where adding noise to the data is equivalent to increasing the classification margin as shown by \citep{xu2009robustness}.\\
As a final remark, it is important to select a proper noise distribution in order to avoid introducing bias and respect the original structure of the data.

\clearpage

\subsection{Experimental parameters} 


\begin{table}[H]
\centering
\resizebox{0.4\textwidth}{!}{
\begin{tabular}{l|c}
 \cline{1-2}
            \multicolumn{2}{c}{Algorithm parameters}         \\  \hline
                      & acGAN \\  \cline{1-2}
\multicolumn{1}{c|}{ \textit{$\alpha$}}    & $0.01$       \\ 
\multicolumn{1}{c|}{ \textit{$\lambda$}}           & $15$    \\ 
\multicolumn{1}{c|}{ \textit{$T_{warmup}$ }}     &  $15 \times N$  \\  
\multicolumn{1}{c|}{ \textit{$N$ }}     & number of discriminators  \\   \hline
  \multicolumn{2}{c}{Optimizer parameters} \\  \hline
\multicolumn{1}{c|}{ \textit{Stacked-MNIST}}     & \multicolumn{1}{|c}{RMSprop ($\alpha=0.9,l_{r}=0.0001$)}  \\ 
\multicolumn{1}{c|}{ \textit{CIFAR-10}}     & \multicolumn{1}{|c}{Adam ($\beta_{1}=0.5,\beta_{2}=0.999,l_{r}=0.0002$)}  \\ 
\multicolumn{1}{c|}{ \textit{Synthetic (25 Gaussians) }}     & \multicolumn{1}{|c}{Adam ($\beta_{1}=0.5,\beta_{2}=0.999,l_{r}=0.0002$)}  \\ 
\multicolumn{1}{c|}{ \textit{Synthetic (8 Gaussians) }}     & \multicolumn{1}{|c}{Adam ($\beta_{1}=0.5,\beta_{2}=0.999,l_{r}=0.0001$)}  \\ 
\multicolumn{1}{c|}{ \textit{CelebA}}     & \multicolumn{1}{|c}{Adam ($\beta_{1}=0.5,\beta_{2}=0.999,l_{r}=0.0002$)}  \\ \hline
\end{tabular}}
\caption{General experimental hyperparameters.}
\label{tab:experiments_parameters}
\end{table}

\subsection{Synthetic data} 


We utilize the 2D-ring with 8 Gaussians and the 2D-grid with 25 Gaussians \cite{wgan-gp}. Three metrics were employed to evaluate the results: 
\begin{enumerate}[leftmargin=*]
    \item $\%$ High Quality samples
    \item Number of Covered modes
    \item Fréchet Distance (FD)
\end{enumerate}

The percentage of "High Quality" samples is defined as the proportion of generated samples $G(z)$ which are within $3$ standard deviation of the closest mode. The next metric reported is the number of modes covered, i.e. the count of modes that has generated samples closes enough ($3 \sigma$). The Fréchet Distance originally from \cite{frechet_original_paper} is defined as:

\begin{equation}
\text{FD} = || m_d - m_g ||^2 + \text{Tr}(C_d+C_g-2(C_d C_g)^{\frac{1}{2}}),
\label{eq:fid_def}
\end{equation}

where $m_d, C_d$ and $m_g, C_g$ are first and second order moments of the real data distributions and estimates from generated data, respectively.

\paragraph{Architecture.}
The generator network's architecture comprises 4 dense layers of $512$ units each. We used 3 discriminators with respectively 2,3 and 4 dense layers of $512$ units. ReLU activations were used in all layers, except for the last one, where a linear activation function was used for the generator and a sigmoid for the discriminator.

\clearpage

\subsection*{Stacked-MNIST} 


\paragraph{Architecture.}

We used DCGAN's architecture \cite{DCGAN} to create lower capacity discriminators (in terms of feature representation power). For the 3Ds case, we used discriminators 3, 4, 5 (described in the following tables). For the 5Ds case, we used discriminators 1, 2, 3, 4, 5.

\begin{table}[H]
\centering
\begin{tabular}{c|c|c|c|c|c}
\hline
Layer                             & Outputs  & Kernel size & Stride & BN & Activation \\ \hline
\multicolumn{1}{c|}{ \textit{Input: $z \sim \mathcal{N}(0, I_{100})$ }}  &          &             &        &            \\
\multicolumn{1}{c|}{ \textit{Fully connected }}                   & 2*2*512  & 4, 4        & 2, 2   & Yes & ReLU       \\
\multicolumn{1}{c|}{ \textit{Transposed convolution }}            & 4*4*256  & 4, 4        & 2, 2   & Yes & ReLU       \\
\multicolumn{1}{c|}{ \textit{Transposed convolution }}            & 8*8*128  & 4, 4        & 2, 2   & Yes  & ReLU       \\
\multicolumn{1}{c|}{ \textit{Transposed convolution }}            & 14*14*64 & 4, 4        & 2, 2   & Yes & ReLU       \\
\multicolumn{1}{c|}{ \textit{Transposed convolution }}            & 28*28*3  & 4, 4        & 2, 2   & No & Tanh       \\ \hline
\end{tabular}
\caption{Generator's architecture.}
\end{table}

\begin{table}[H]
\centering
\begin{tabular}{c|c|c|c|c|c}
\hline
Layer                             & Outputs  & Kernel size & Stride & BN & Activation \\ \hline
\multicolumn{1}{c|}{ \textit{Input }}  &    28*28*3      &             &        &            \\
\multicolumn{1}{c|}{ \textit{ Convolution }}                   & 14*14*64  & 4, 4        & 2, 2   & No &     LeakyReLU    \\
\multicolumn{1}{c|}{ \textit{ Convolution }}            & 7*7*128  & 4, 4        & 2, 2   & Yes & LeakyReLU       \\
\multicolumn{1}{c|}{ \textit{ Convolution }}            & 4*4*256  & 4, 4        & 2, 2   & Yes  & LeakyReLU       \\
\multicolumn{1}{c|}{ \textit{ Convolution }}            & 2*2*512 & 4, 4        & 2, 2   & Yes & LeakyReLU       \\
\multicolumn{1}{c|}{ \textit{ Convolution }}            & 1  & 4, 4        & 2, 2   & No & Sigmoid       \\ \hline
\end{tabular}
\caption{Discriminator 5.}
\end{table}

\begin{table}[H]
\centering
\begin{tabular}{c|c|c|c|c|c}
\hline
Layer                             & Outputs  & Kernel size & Stride & BN & Activation \\ \hline
\multicolumn{1}{c|}{ \textit{Input }}  &   28*28*3       &             &        &            \\
\multicolumn{1}{c|}{ \textit{ Convolution }}                   & 14*14*64  & 4, 4        & 2, 2   & No &    LeakyReLU     \\
\multicolumn{1}{c|}{ \textit{ Convolution }}            & 7*7*128  & 4, 4        & 2, 2   & Yes & LeakyReLU       \\
\multicolumn{1}{c|}{ \textit{ Convolution }}            & 4*4*256  & 4, 4        & 2, 2   & Yes  & LeakyReLU       \\
\multicolumn{1}{c|}{ \textit{ Convolution }}            & 2*2*512 & 4, 4        & 2, 2   & Yes & LeakyReLU       \\
\multicolumn{1}{c|}{ \textit{ Fully connected }}            & 1  & 4, 4        & 2, 2   & No & Sigmoid       \\ \hline
\end{tabular}
\caption{Discriminator 4.}
\end{table}

\begin{table}[H]
\centering
\begin{tabular}{c|c|c|c|c|c}
\hline
Layer                             & Outputs  & Kernel size & Stride & BN & Activation \\ \hline
\multicolumn{1}{c|}{ \textit{Input }}  &   28*28*3       &             &        &            \\
\multicolumn{1}{c|}{ \textit{ Convolution }}                   & 13*13*64  & 6, 6        & 2, 2   & No &   LeakyReLU      \\
\multicolumn{1}{c|}{ \textit{ Convolution }}            & 6*6*128  & 6, 6        & 2, 2   & Yes & LeakyReLU       \\
\multicolumn{1}{c|}{ \textit{ Convolution }}            & 2*2*256  & 6, 6        & 2, 2   & Yes  & LeakyReLU       \\
\multicolumn{1}{c|}{ \textit{ Convolution }}            & 1  & 6, 6        & 2, 2   & No & Sigmoid       \\ \hline
\end{tabular}
\caption{Discriminator 3.}
\end{table}

\begin{table}[H]
\centering
\begin{tabular}{c|c|c|c|c|c}
\hline
Layer                             & Outputs  & Kernel size & Stride & BN & Activation \\ \hline
\multicolumn{1}{c|}{ \textit{Input }}  &   28*28*3       &             &        &            \\
\multicolumn{1}{c|}{ \textit{ Convolution }}                   & 13*13*64  & 6, 6        & 2, 2   & No &   LeakyReLU      \\
\multicolumn{1}{c|}{ \textit{ Convolution }}            & 6*6*128  & 6, 6        & 2, 2   & Yes & LeakyReLU       \\
\multicolumn{1}{c|}{ \textit{ Convolution }}            & 2*2*256  & 6, 6        & 2, 2   & Yes  & LeakyReLU       \\
\multicolumn{1}{c|}{ \textit{ Fully connected }}            & 1  & 6, 6        & 2, 2   & No & Sigmoid       \\ \hline
\end{tabular}
\caption{Discriminator 2.}
\end{table}

\begin{table}[H]
\centering
\begin{tabular}{c|c|c|c|c|c}
\hline
Layer                             & Outputs  & Kernel size & Stride & BN & Activation \\ \hline
\multicolumn{1}{c|}{ \textit{Input }}  &   28*28*3       &             &        &            \\
\multicolumn{1}{c|}{ \textit{ Convolution }}                   & 12*12*64  & 8, 8        & 2, 2   & No &  LeakyReLU       \\
\multicolumn{1}{c|}{ \textit{ Convolution }}            & 4*4*128  & 8, 8        & 2, 2   & Yes & LeakyReLU       \\
\multicolumn{1}{c|}{ \textit{ Convolution }}            & 1 & 8, 8        & 2, 2   & No & Sigmoid       \\  \hline
\end{tabular}
\caption{Discriminator 1.}
\end{table}

\begin{figure}[H]
    \begin{subfigure}[htbp]{0.5\textwidth}
        \centering
        \includegraphics[width=6.5cm]{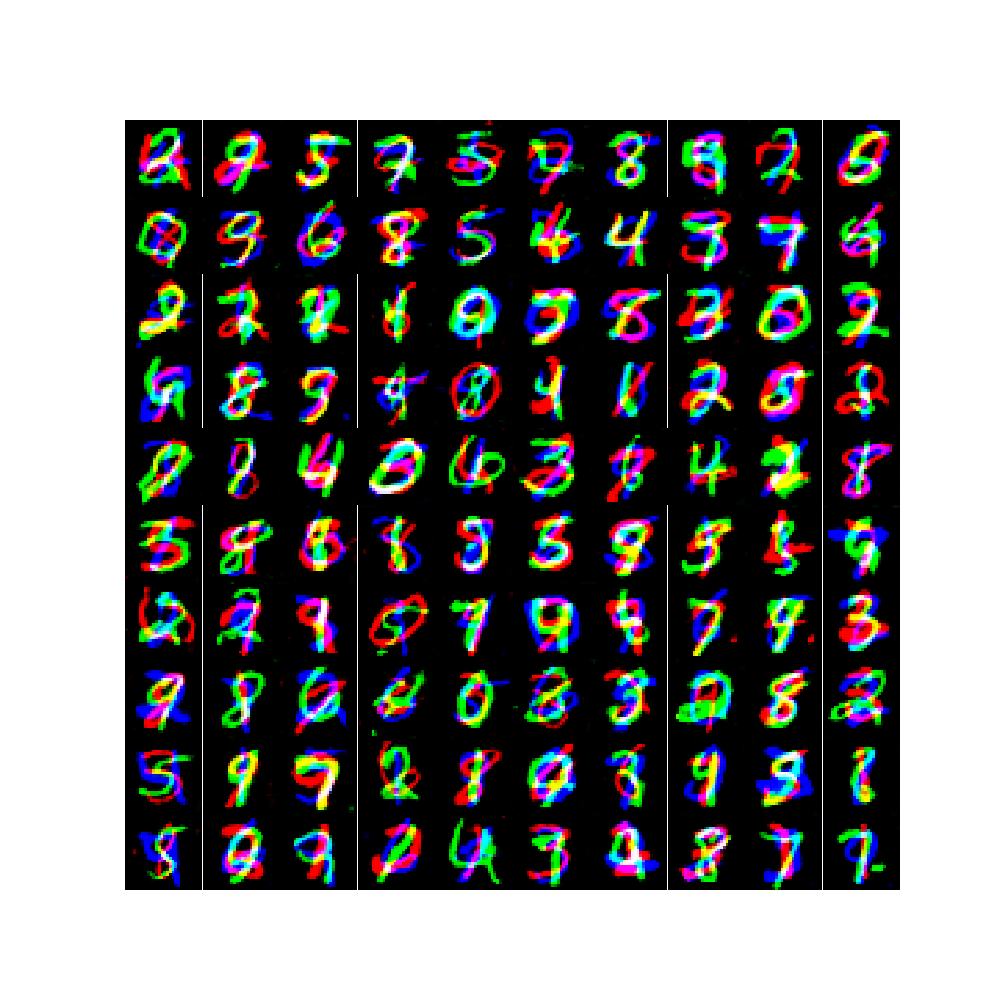}
        \caption{acGAN - 3 disc.}
    \end{subfigure}%
    ~ 
    \begin{subfigure}[htbp]{0.5\textwidth}
        \centering
        \includegraphics[width=6.5cm]{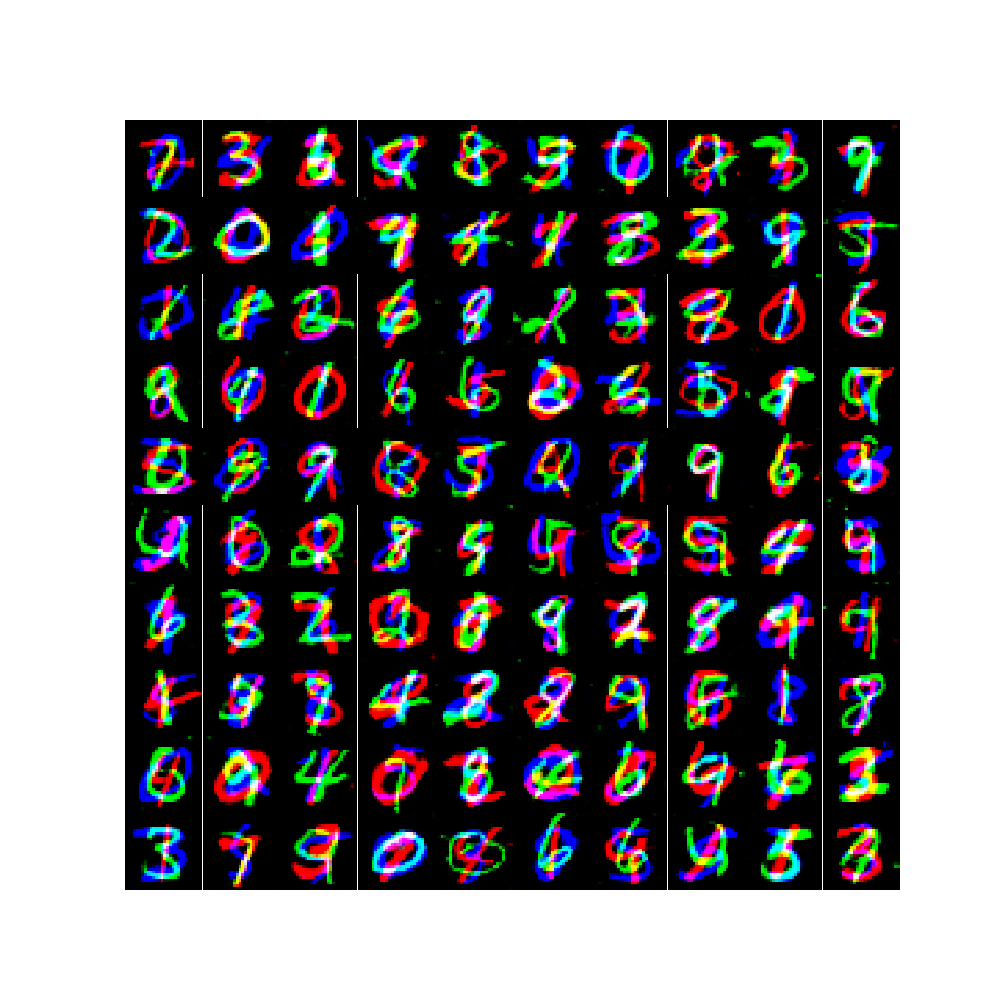}
        \caption{acGAN - 5 disc.}
    \end{subfigure}
    \caption{Stacked-MNIST generated samples.}
    
\end{figure}
\clearpage

\subsection{CIFAR-10} 


\noindent
\textbf{FID score.} FID scores, as introduced in \cite{FID}, were computed for CIFAR-10. It is defined as the squared Fr\'echet distance between the Gaussian having the first and second order statistics matching those obtained from image features. The late layers of a pretrained classifier  are used as low dimensional representation of images for statistics estimation. 

\noindent
\textbf{Architecture.} For our strongest discriminator we use the DCGAN architecture but with halved the number of filter, i.e. $\{64,128,256,512\}$. For the 3D case, we introduced two extra discriminators with kernel sizes of 6 and 8. For the 5D case, we add  two discriminators with kernel sizes 4 and 6 respectively to the set of 3D discriminator networks. In both 3D and 5D, we replaced the last layer from the DCGAN model with a fully connected dense layer. The generator network was taken from the original DCGAN architecture but with halved filter sizes too, i.e. $\{512,256,128,64\}$. ReLU activation units were used for the generator network while LeakyRelu is used for the discriminators with a coefficient of $0.2$. 

\begin{figure}[H]
    \centering
   \includegraphics[scale=0.49]{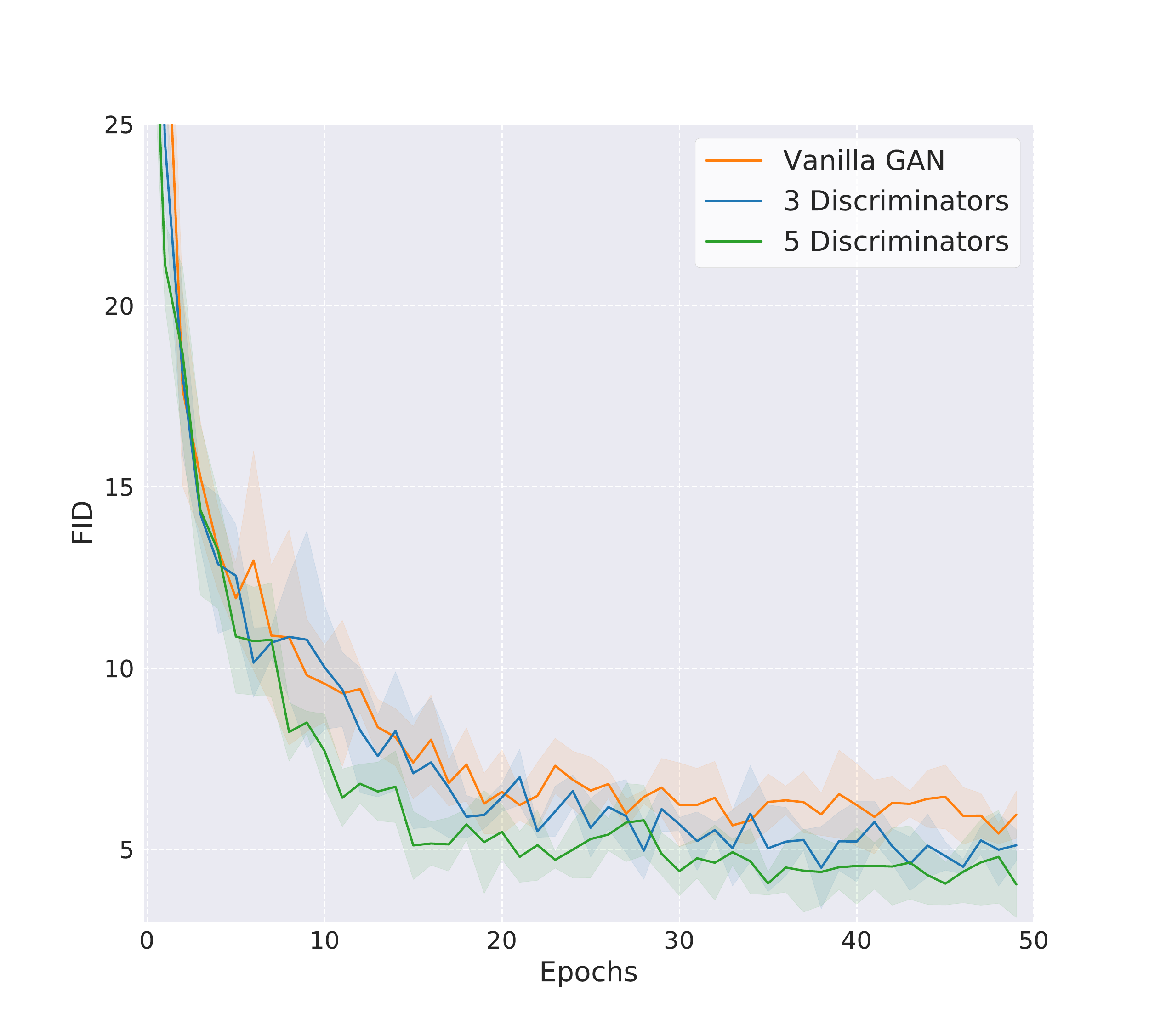}
   \centering
    \caption{ Increasing the number of discriminators induces an earlier convergence of FID. Moreover, lower FID values are reached.}
    \label{fig:influence_nb_disc}
\end{figure}

\noindent
\textbf{Influence of the number of discriminators.} An important assumption in the current paper is that increasing the number of discriminator networks helps the model converge faster. To assess that, we conducted experiments with the acGAN algorithm while varying the number of discriminators for $N \in \{1,3,5\}$ ($N=1$ being the Vanilla GAN) and averaging results over 5 seeds. According to Fig. \ref{fig:influence_nb_disc}, we see that a higher number of discriminators indeed leads to earlier convergence of the FID score curve.

\begin{figure}[H]
    \centering
    \begin{subfigure}[b]{\columnwidth}        
        \centering
        \includegraphics[width=12cm,height=9cm]{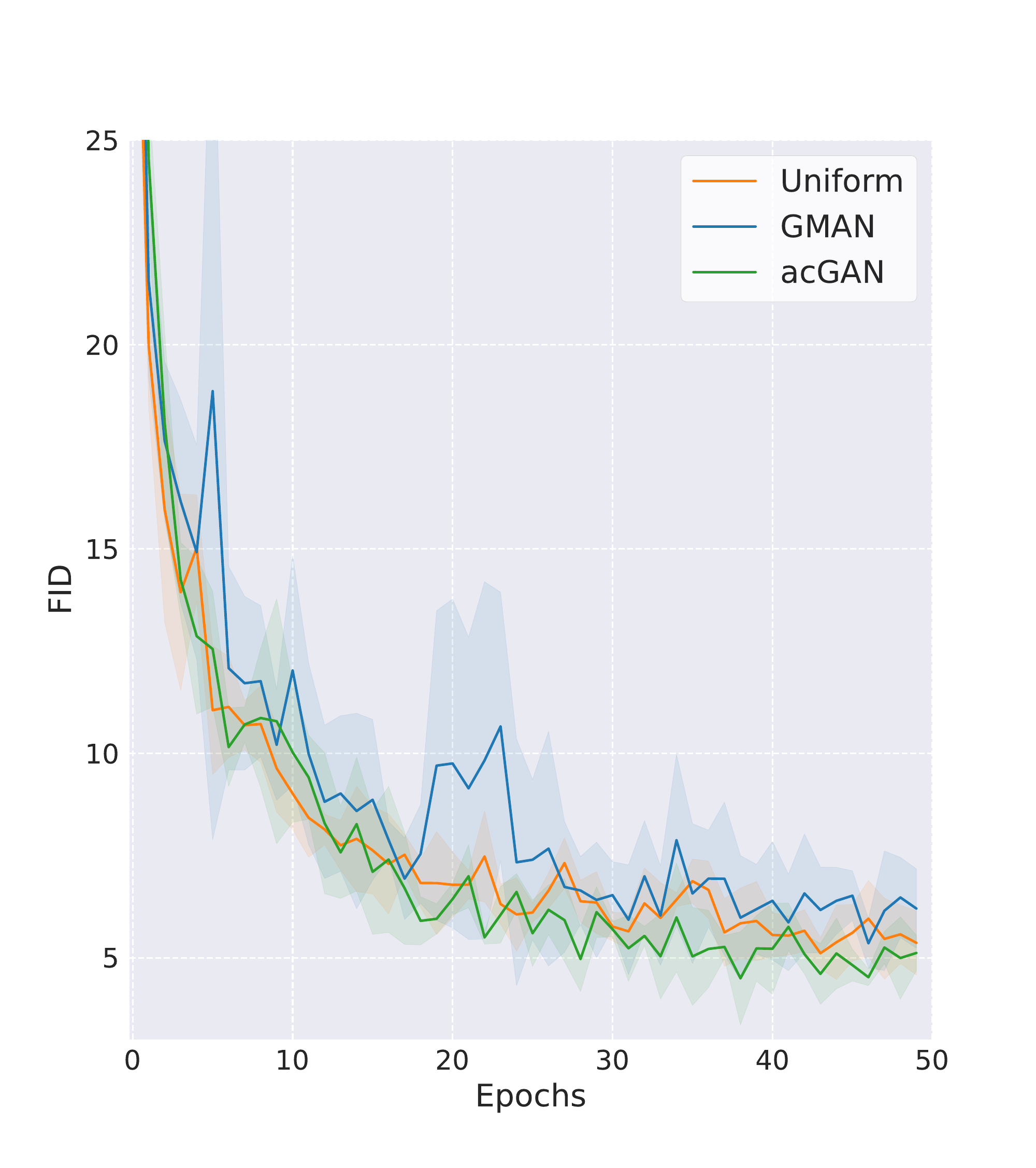}
        \caption{3 Discriminators}
        \label{fig:A}
    \end{subfigure}
    \hfill
    \begin{subfigure}[b]{\columnwidth}        
        \centering
        \includegraphics[width=12cm,height=9cm]{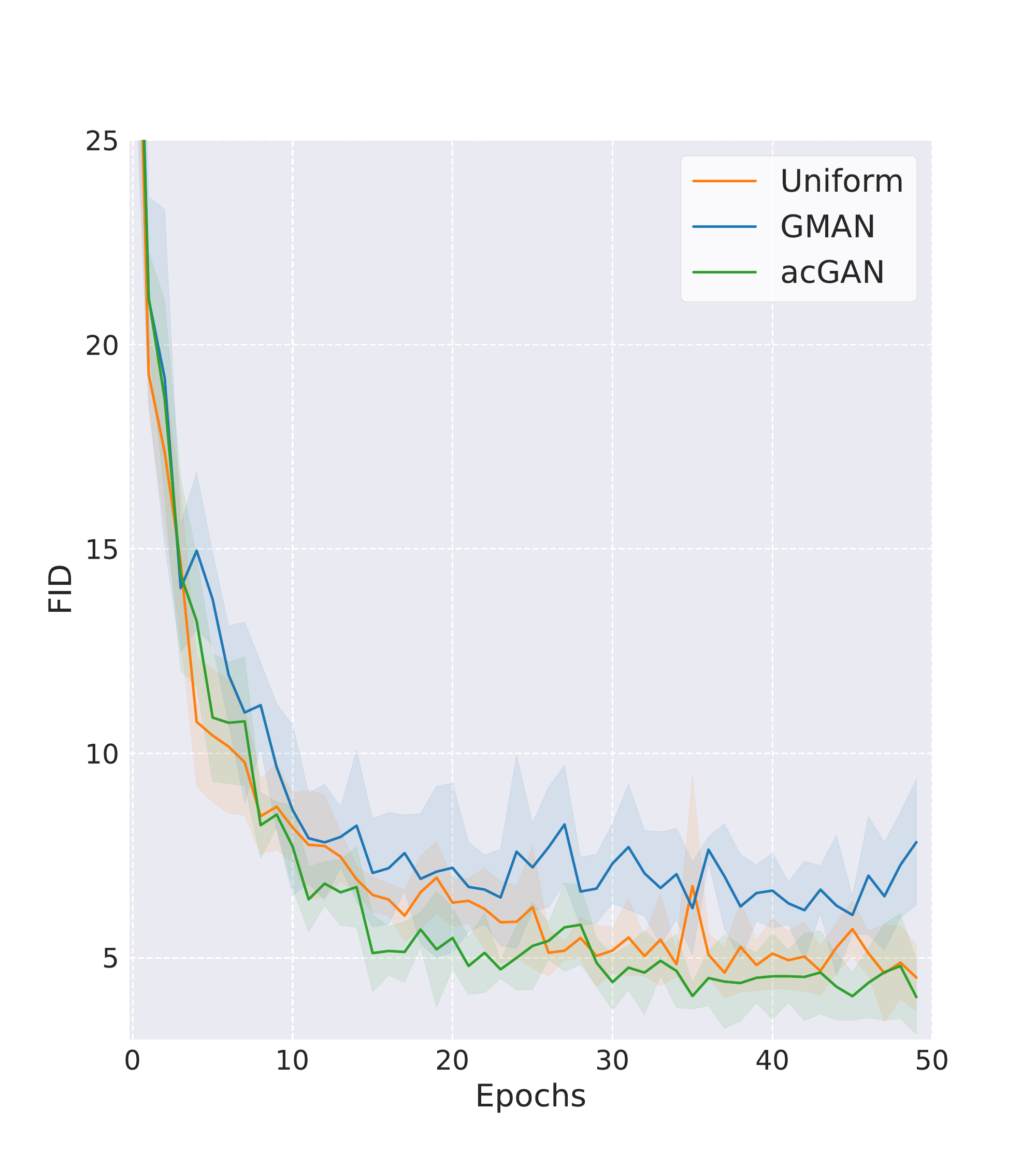}
        \caption{5 Discriminators}
        \label{fig:B}
    \end{subfigure}
    \caption{Average FID score of each method for different number of discriminators. In both plots, the acGAN algorithm presented faster convergence compared to the other methods.}
    \label{fig:comparison_all_cifar10}
\end{figure}

\begin{figure}[H]
    \centering
   \includegraphics[scale=0.49]{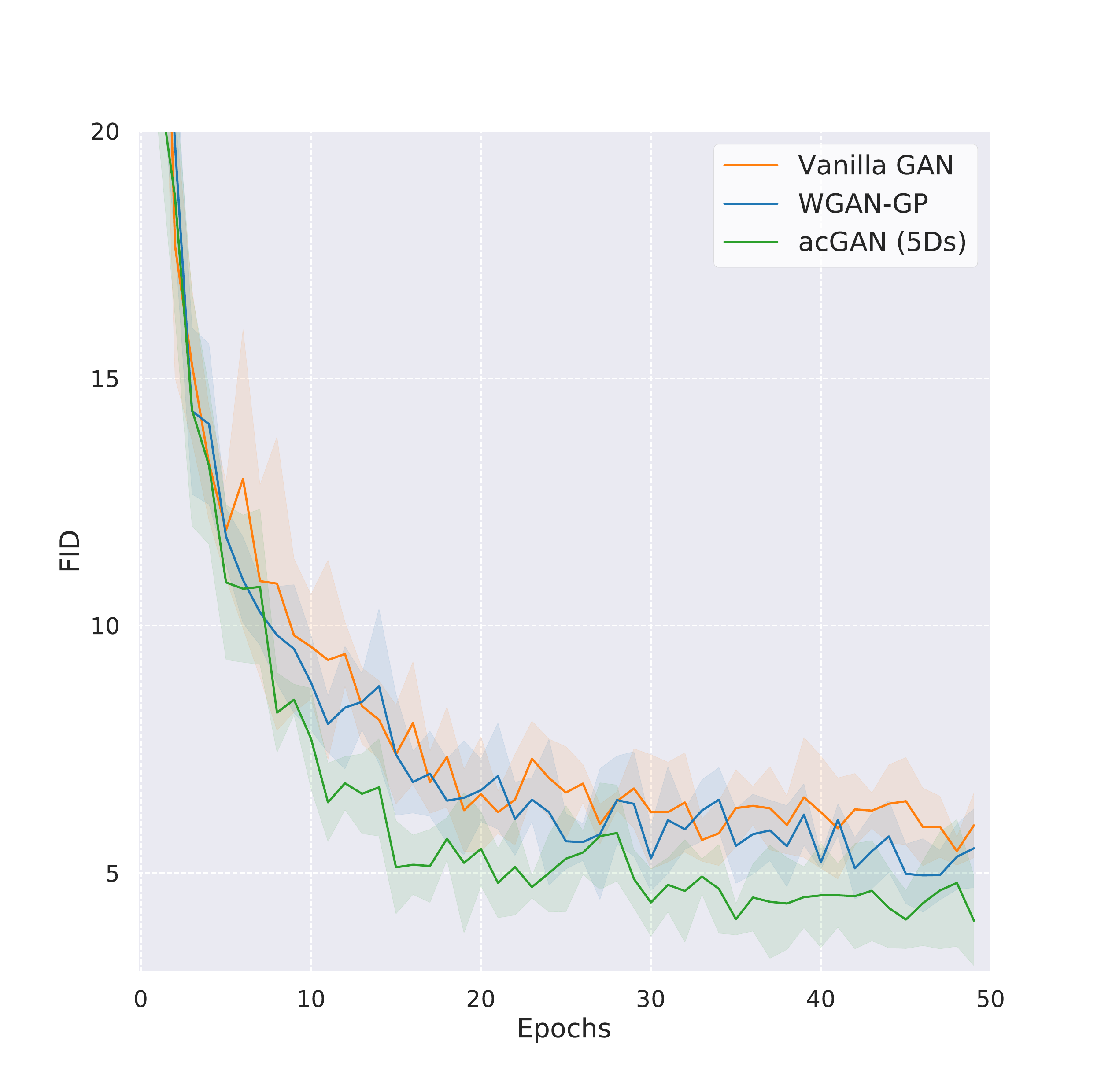}
   \centering
    \caption{ acGAN with 5 discriminators shows earler convergence and better performance than Vanilla GAN (1 Disc) and WGAN-GP.}
    \label{fig:comparison_wgan_acgan}
\end{figure}

\clearpage

\vspace{10cm}
\begin{figure}[H]
   \begin{subfigure}[htbp]{0.5\textwidth}
        \centering
        \includegraphics[width=6.5cm]{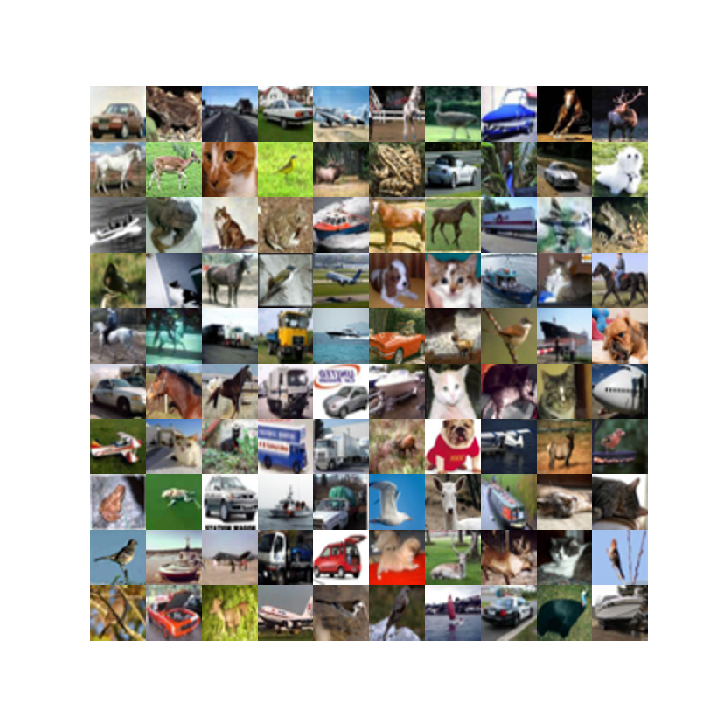}
        \caption{Real Images}
    \end{subfigure}%
    ~ 
    \begin{subfigure}[htbp]{0.5\textwidth}
        \centering
        \includegraphics[width=6.5cm]{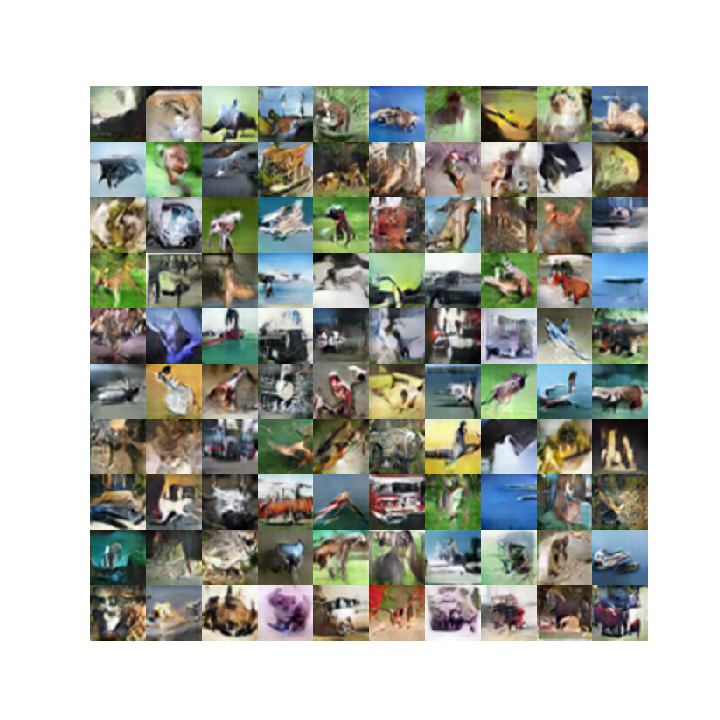}
        \caption{Vanilla GAN}
    \end{subfigure}
	 ~ 
    \begin{subfigure}[htbp]{0.5\textwidth}
        \centering
        \includegraphics[width=6.5cm]{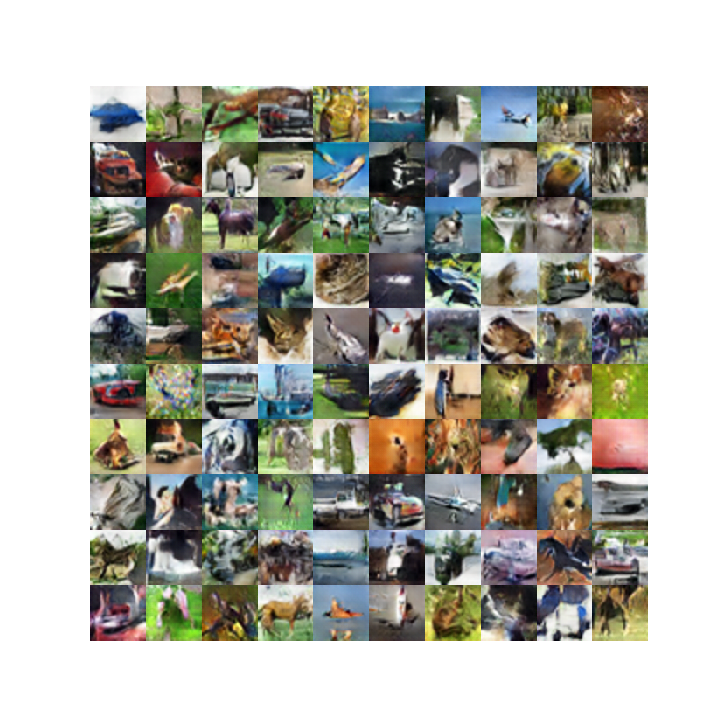}
        \caption{Uniform - 3 disc.}
    \end{subfigure}%
    ~ 
    \begin{subfigure}[htbp]{0.5\textwidth}
        \centering
        \includegraphics[width=6.5cm]{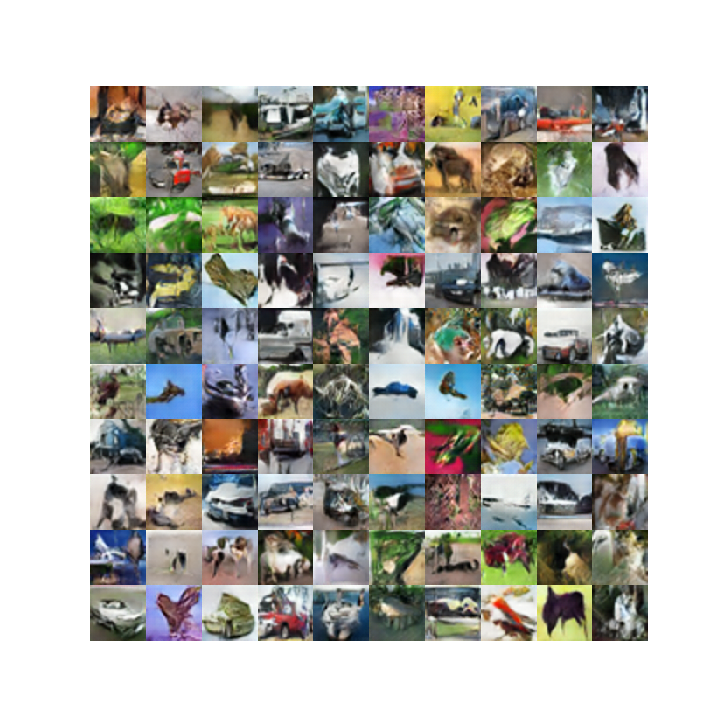}
        \caption{Uniform - 5 disc.}
    \end{subfigure}
	 ~ 
        \begin{subfigure}[htbp]{0.5\textwidth}
        \centering
        \includegraphics[width=6.5cm]{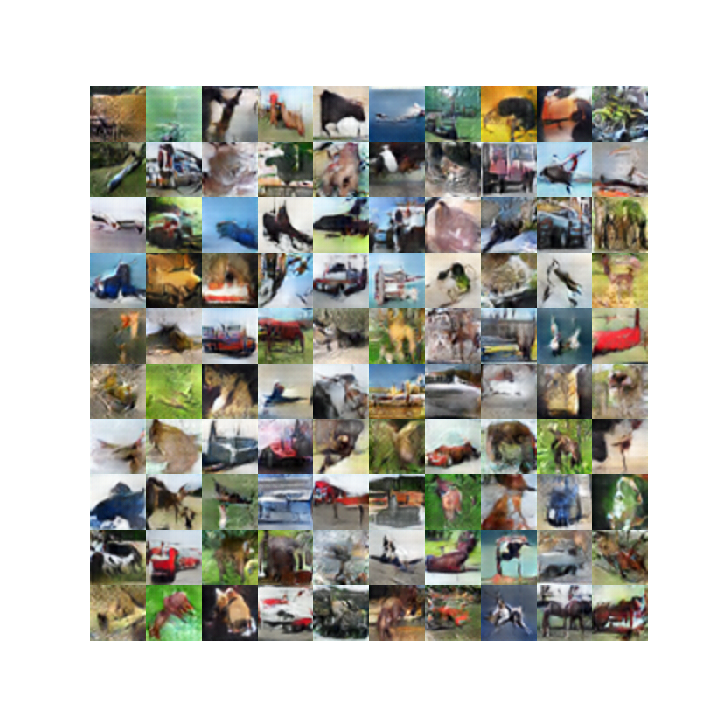}
        \caption{GMAN - 3 disc.}
    \end{subfigure}%
    ~ 
    \begin{subfigure}[htbp]{0.5\textwidth}
        \centering
        \includegraphics[width=6.5cm]{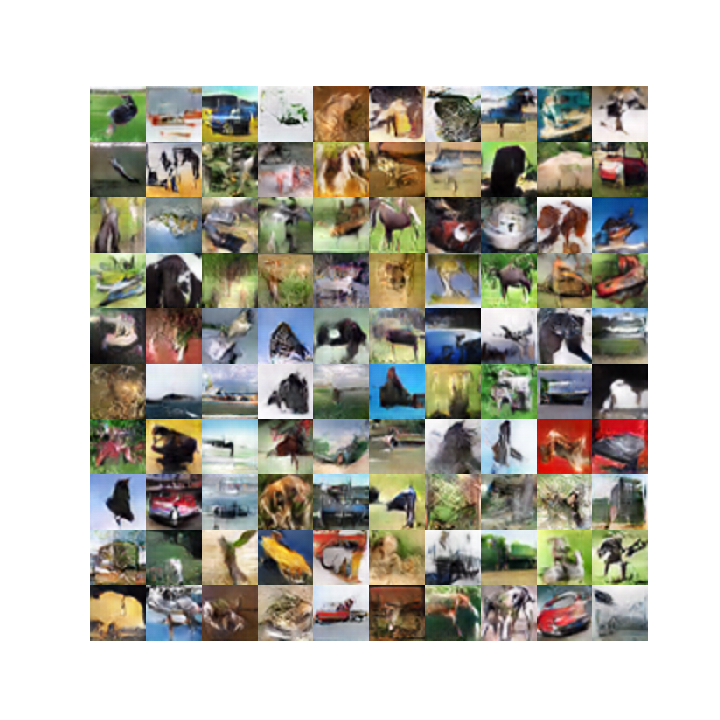}
        \caption{GMAN - 5 disc.}
    \end{subfigure}
     \caption{CIFAR-10 generated samples (1).}
      \label{fig:cifar10_samples}
\end{figure}

\begin{figure}[H]
    \begin{subfigure}[htbp]{0.5\textwidth}
        \centering
        \includegraphics[width=6.5cm]{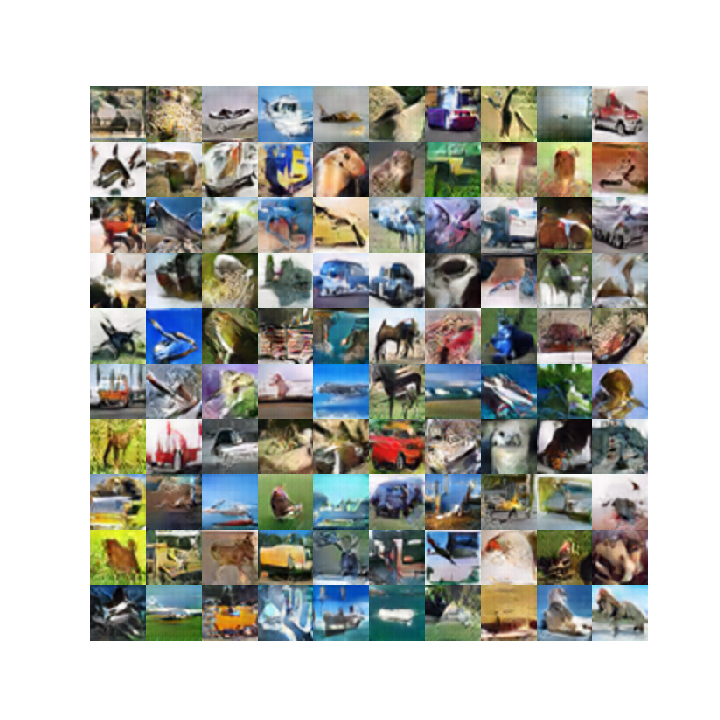}
        \caption{acGAN - 3 disc.}
    \end{subfigure}%
    ~ 
    \begin{subfigure}[htbp]{0.5\textwidth}
        \centering
        \includegraphics[width=6.5cm]{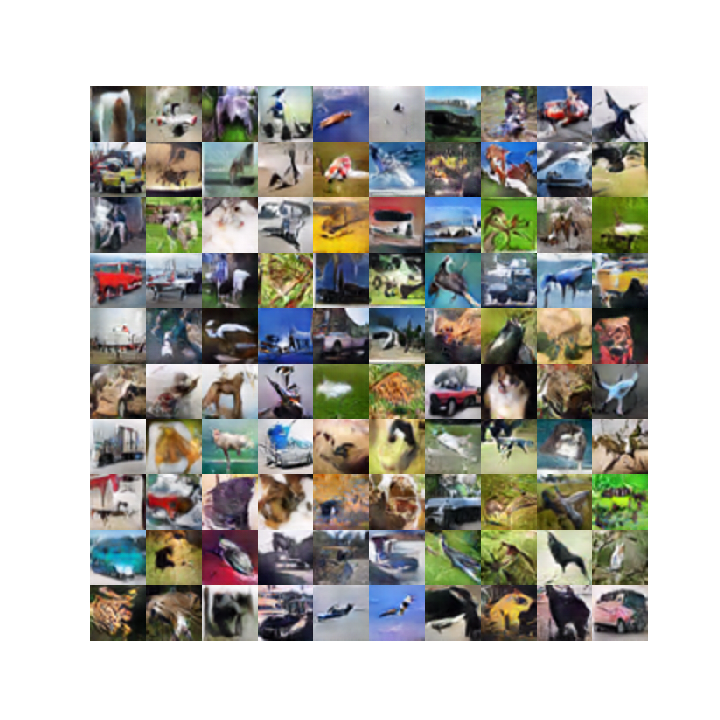}
        \caption{acGAN - 5 disc.}
    \end{subfigure}
    \caption{CIFAR-10 generated samples (2).}
\end{figure}
\clearpage

\subsection*{CelebA} 


For both the CelebA \citep{celebA} datasets, we conducted single-run experiments of 50,000 iterations each counted in generator steps ($\approx 15$ epochs). We downscaled the original images to $64 \times 64$ pixels out of practical concerns.

\paragraph{Results.}

\begin{figure}[H]
    \centering
  \includegraphics[scale=0.6]{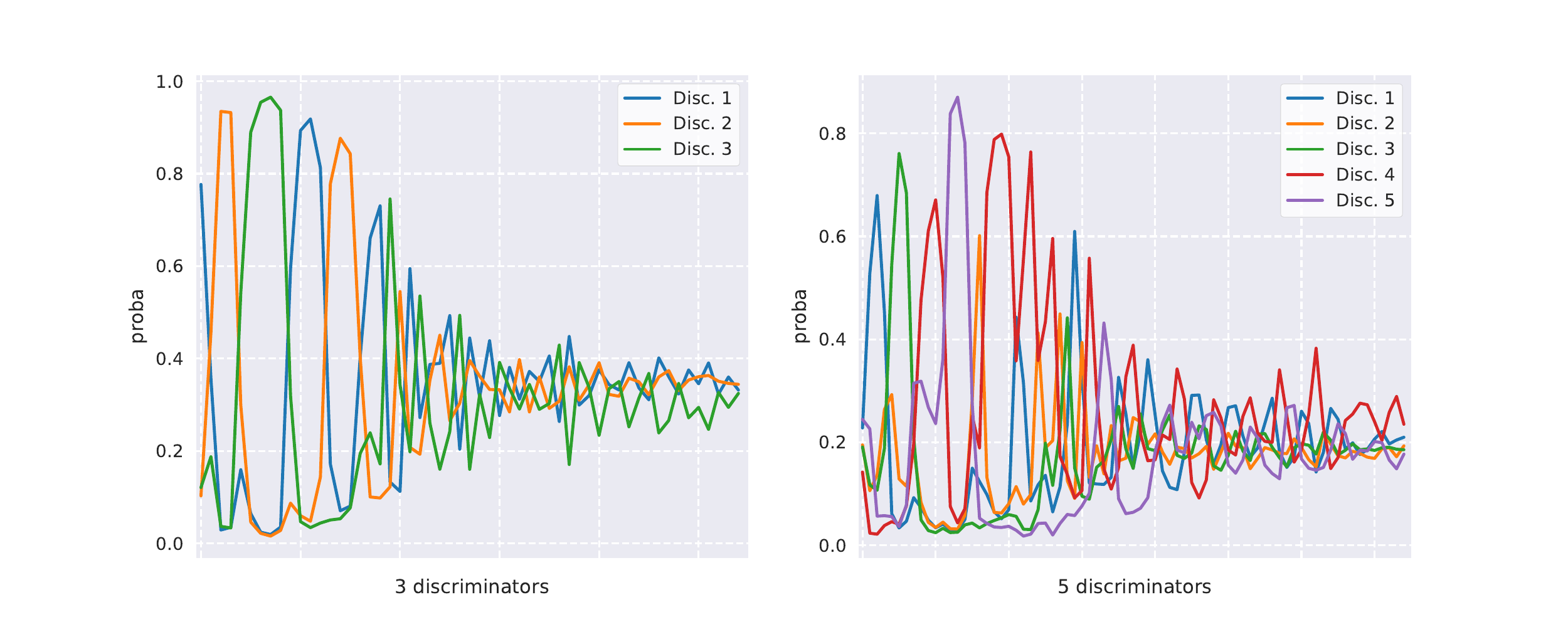}
    \caption{Weight $\pi_{i}$ of each discriminator over the training epochs. We could see switching phase, where one discriminator's weight $\pi_{i}$ is dominant with respect to the rest. After some epochs, all weights $\pi_{i}$ converge to a uniform regime.}
    \label{fig:sampling_proba_celebA}
\end{figure}

Similarly to CIFAR-10, we observe the emergence of a curriculum in Fig. \ref{fig:sampling_proba_celebA}. In particular, we note the presence of alternating phases during which a specific discriminator is dominating in the 3D and 5D cases. In the end, all discriminator probabilities converge to a stationary (\textit{i.e.} long term) uniform distribution just like for previously mentioned datasets.
\\

\paragraph{Architecture.}
We used the same architecture as for the CIFAR-10 experiments except that the original numbers of filters were set to: $\{128,256,512,1024\}$ for the discriminators and $\{1024,512,256,128\}$ for the generator.

\begin{figure}[H]
\begin{subfigure}[htbp]{0.5\textwidth}
        \centering
        \includegraphics[width=6.5cm]{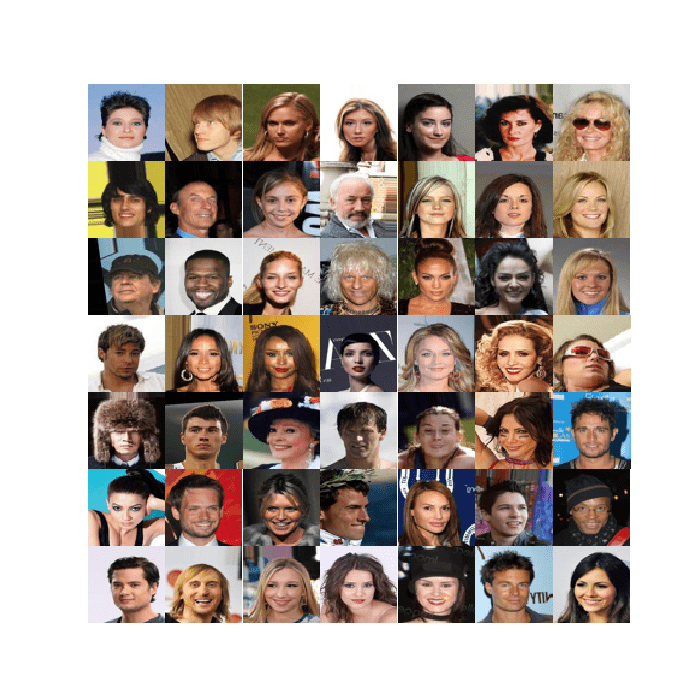}
        \caption{Real Images}
    \end{subfigure}%
    ~ 
    \begin{subfigure}[htbp]{0.5\textwidth}
        \centering
        \includegraphics[width=6.5cm]{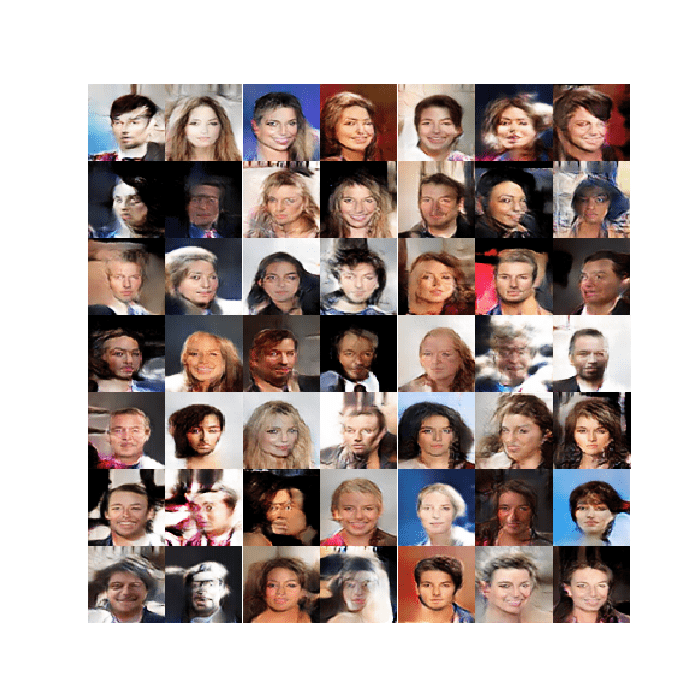}
        \caption{Vanilla GAN}
    \end{subfigure}
    \begin{subfigure}[htbp]{0.5\textwidth}
        \centering
        \includegraphics[width=6.5cm]{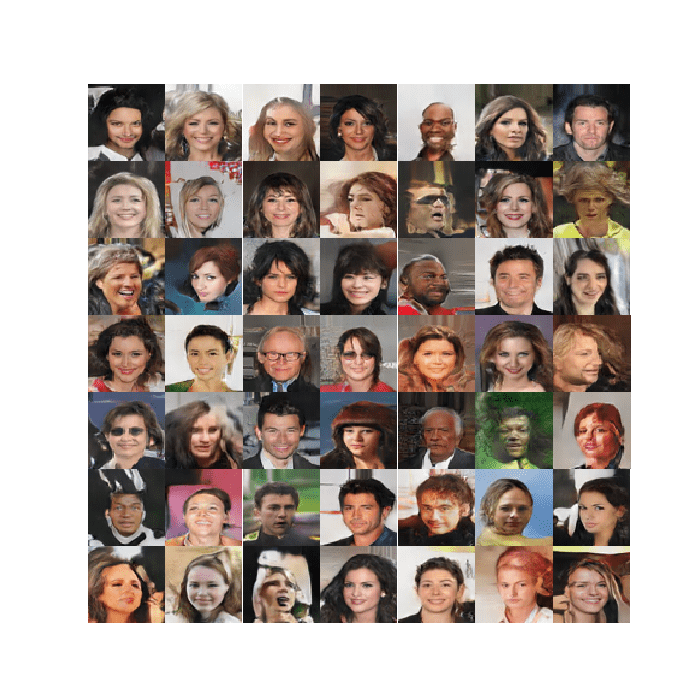}
        \caption{Uniform - 3 disc.}
    \end{subfigure}%
    ~ 
    \begin{subfigure}[htbp]{0.5\textwidth}
        \centering
        \includegraphics[width=6.5cm]{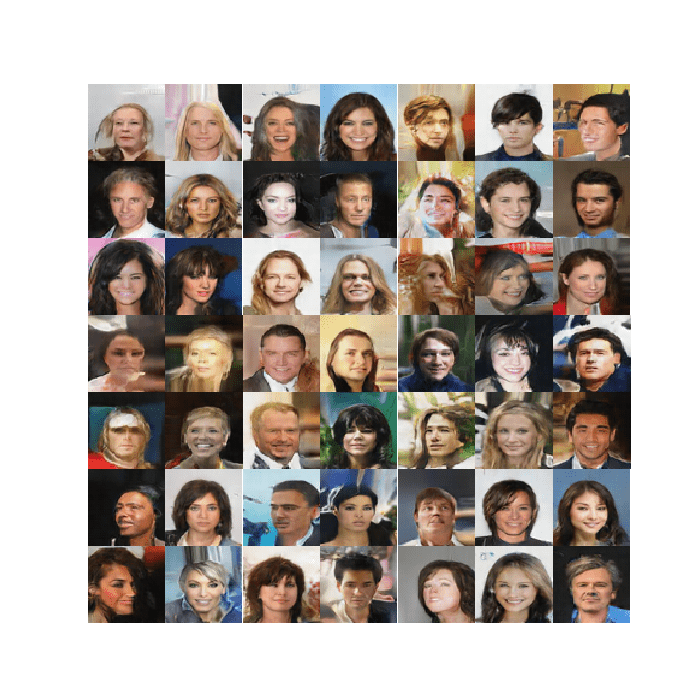}
        \caption{Uniform - 5 disc.}
    \end{subfigure}
    \caption{CelebA generated samples (1).}
     \label{fig:celeba_samples}
\end{figure}

\begin{figure}[H]

     \begin{subfigure}[htbp]{0.5\textwidth}
        \centering
        \includegraphics[width=6.5cm]{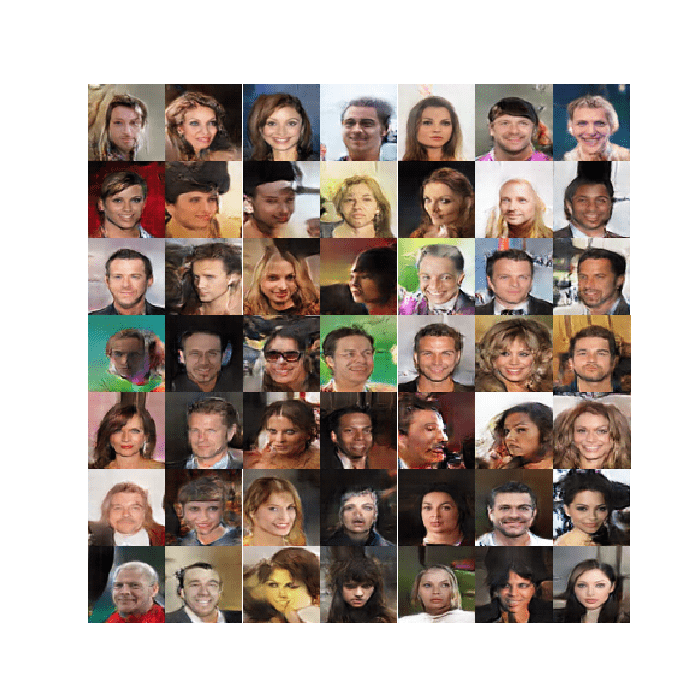}
        \caption{GMAN - 3 disc.}
    \end{subfigure}%
    ~ 
    \begin{subfigure}[htbp]{0.5\textwidth}
        \centering
        \includegraphics[width=6.5cm]{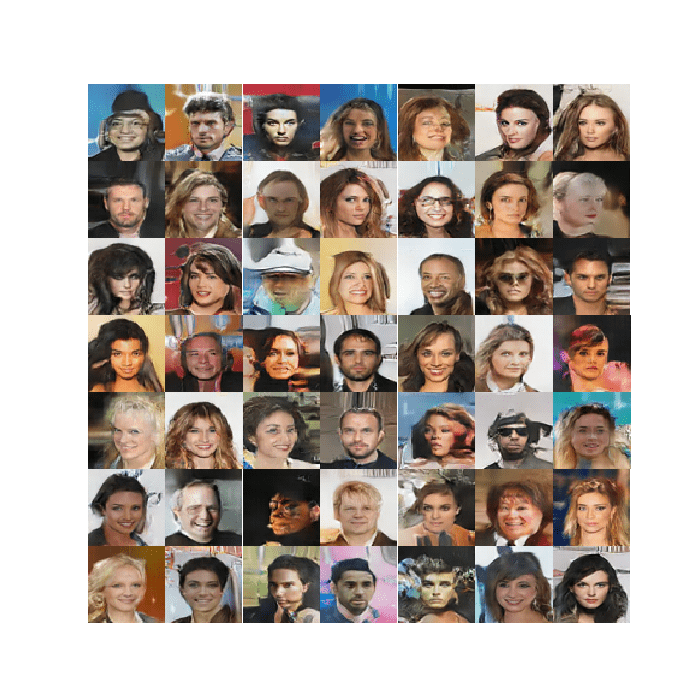}
        \caption{GMAN - 5 disc.}
    \end{subfigure}
    
     ~  
    \begin{subfigure}[h]{0.5\textwidth}
        \centering
        \includegraphics[width=6.5cm]{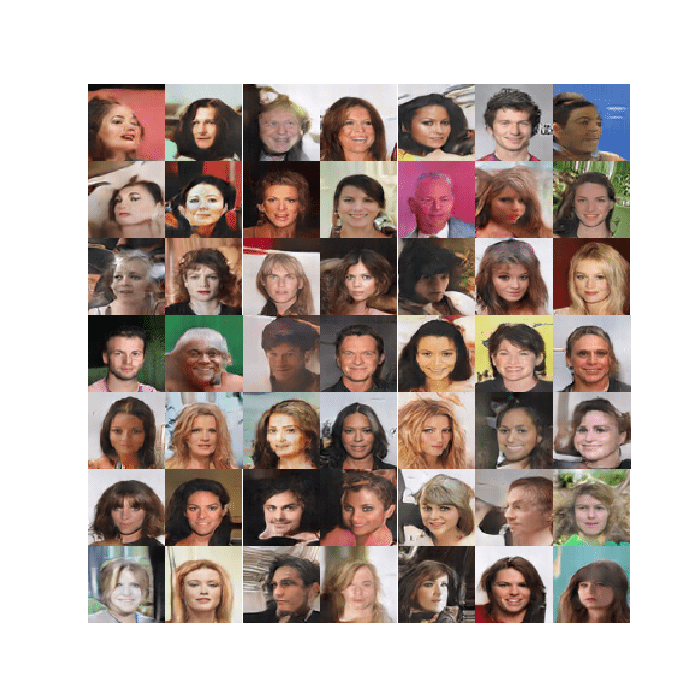}
        \caption{acGAN - 3 disc.}
    \end{subfigure}%
    ~ 
    \begin{subfigure}[h]{0.5\textwidth}
        \centering
        \includegraphics[width=6.5cm]{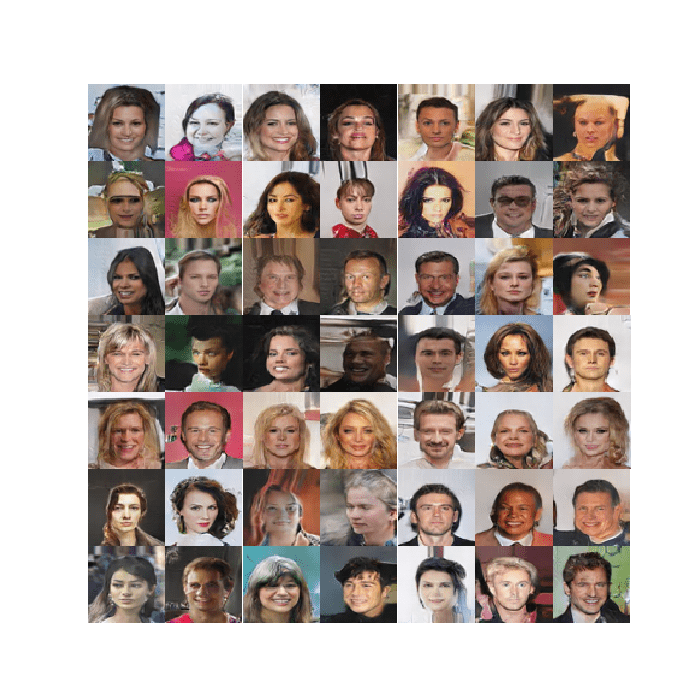}
        \caption{acGAN - 5 disc.}
    \end{subfigure}
    \caption{CelebA generated samples (2).}
    
\end{figure}

\paragraph{Generating 128x128 images.}
In this experiment, we generated high resolution images with 3 and 5 discriminators. A convolutional layer with 2048 feature maps was added to both generator and discriminators architectures. The 3 discriminators settings used a kernel size of 4,6 and 8. For the 5 discriminators case, we added a discriminator of kernel size 4 and 6 but replaced the last layer with dense layers. The same parameters was employed as for CelebA 64x64.

\begin{figure}[H]
    \centering
    \begin{subfigure}[h]{\textwidth}
        \hspace{-0.60in}
        \includegraphics[width = 8.22in]{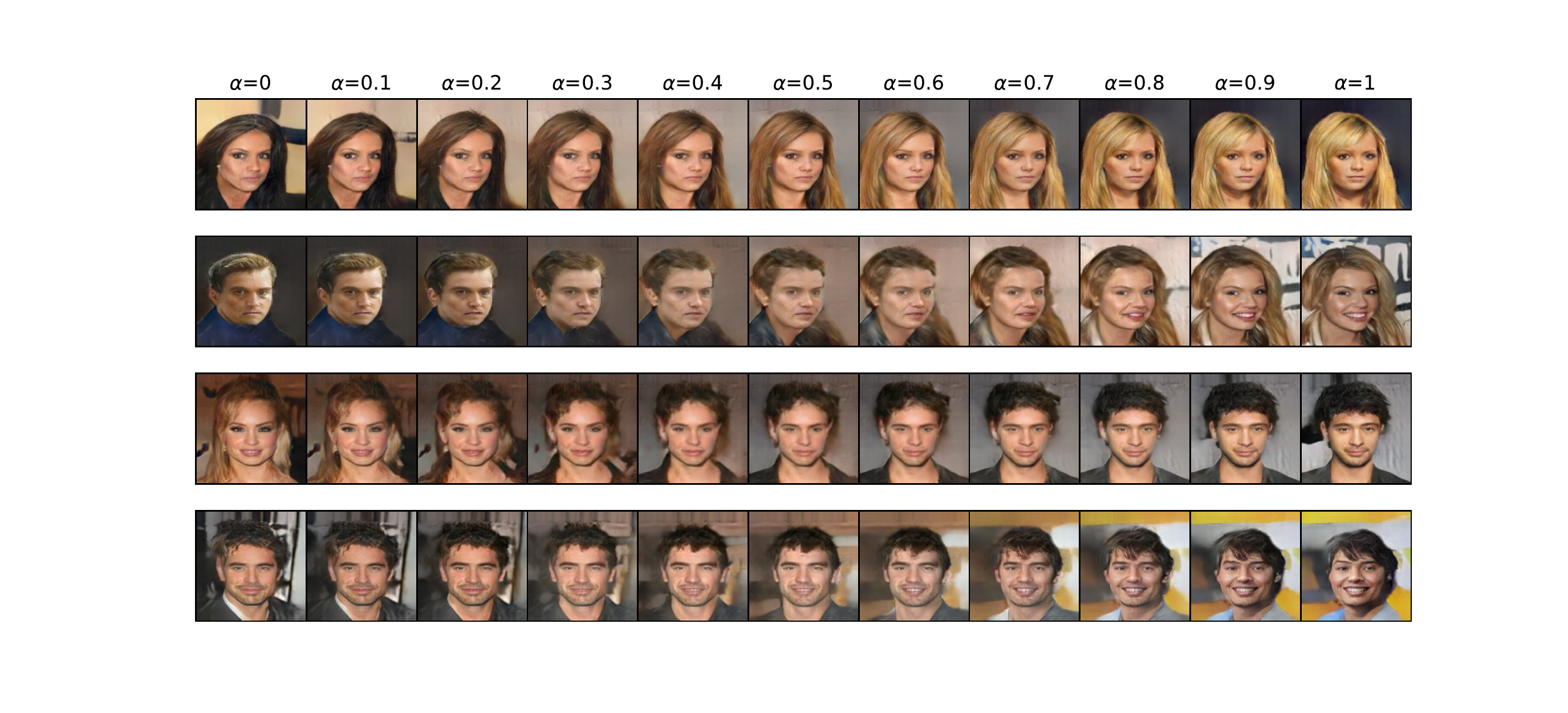}
        \caption{acGAN - 3 discriminators}
    \end{subfigure}%
    
     \begin{subfigure}[h]{\textwidth}
         \hspace{-0.60in}
        \includegraphics[width = 8.22in]{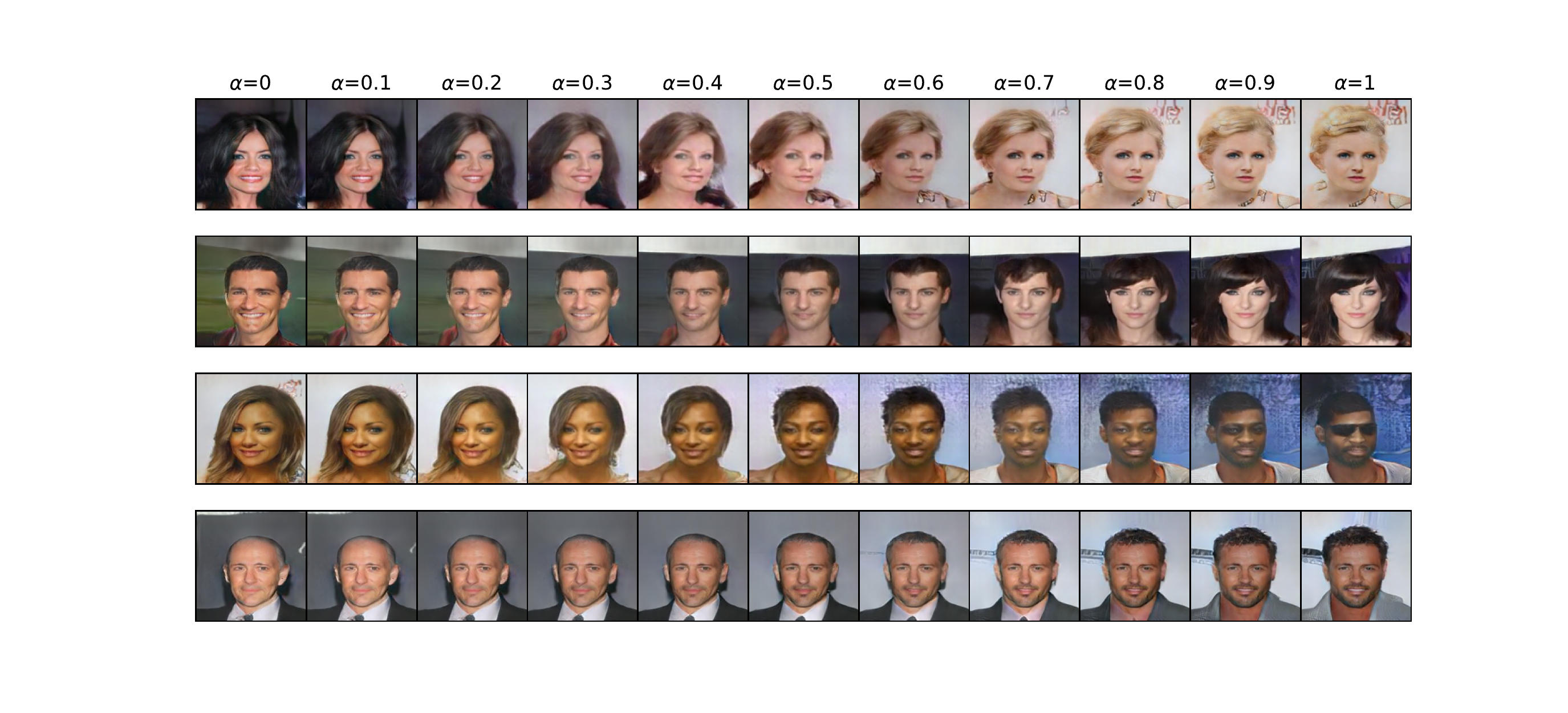}
        \caption{acGAN - 5 discriminators}
    \end{subfigure}%
    \caption{Interpolating in latent space with 3 and 5 Discriminators.} 
     \label{fig:interpolation_celeba}
\end{figure}

\begin{figure}[H]
    \centering
    \begin{subfigure}[H]{\textwidth}
    \centering
    \begin{tabular}{cccc}
    \includegraphics[width = 1.4in]{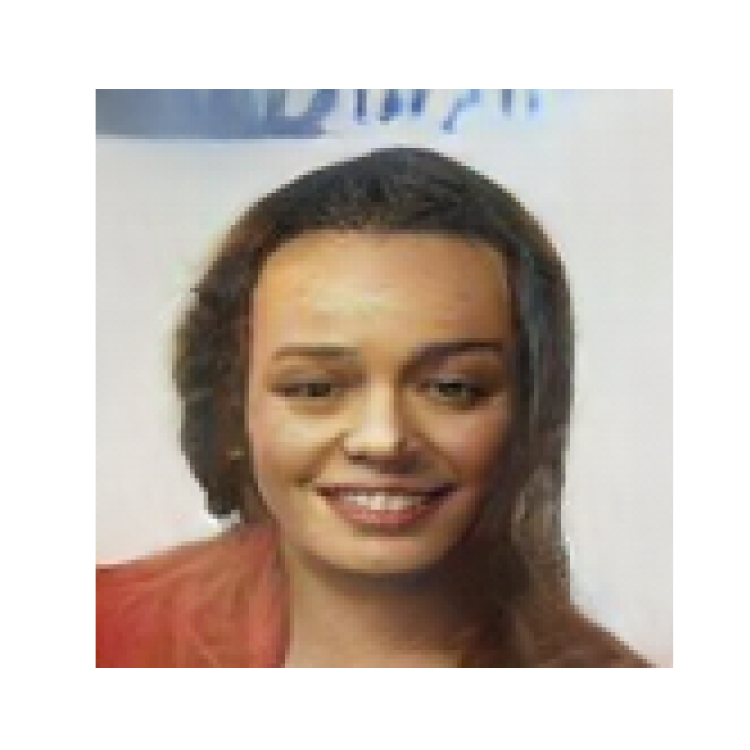}&
    \includegraphics[width = 1.4in]{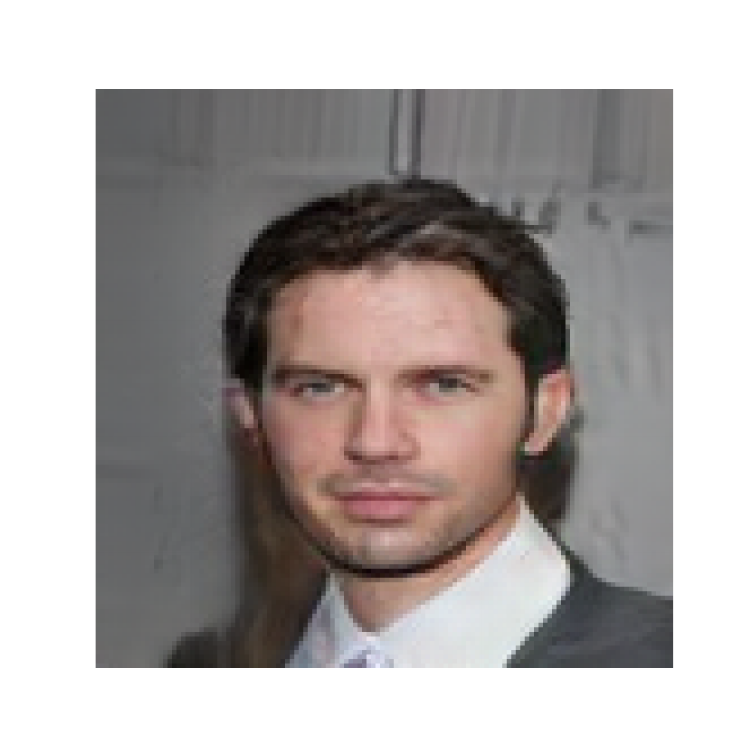} &
    \includegraphics[width = 1.4in]{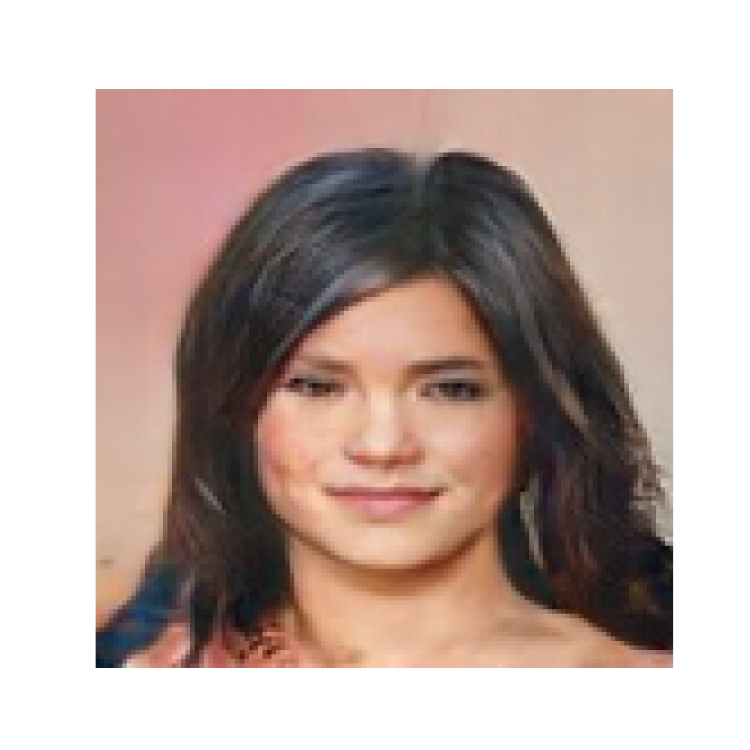} &
    \includegraphics[width = 1.4in]{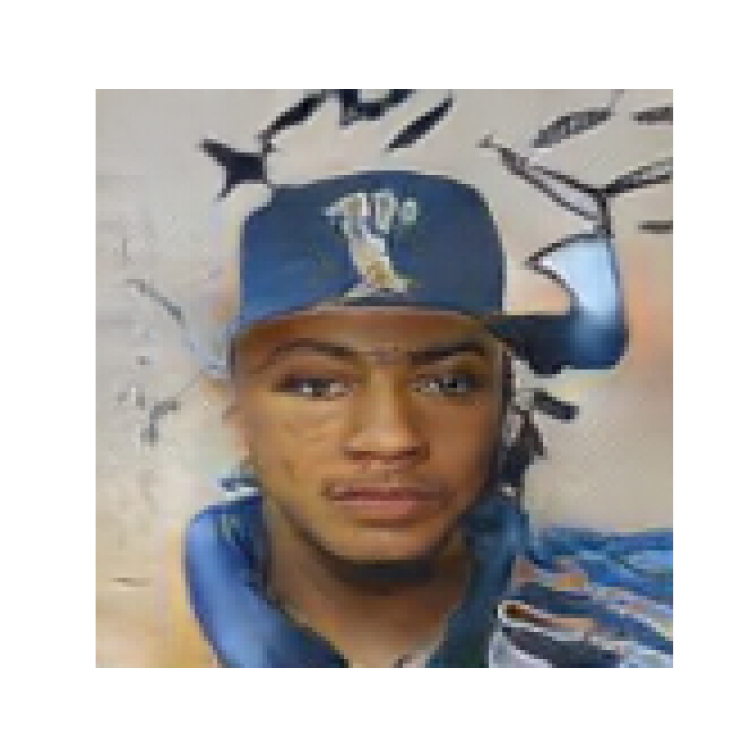} 
    
    \\
    \includegraphics[width = 1.4in]{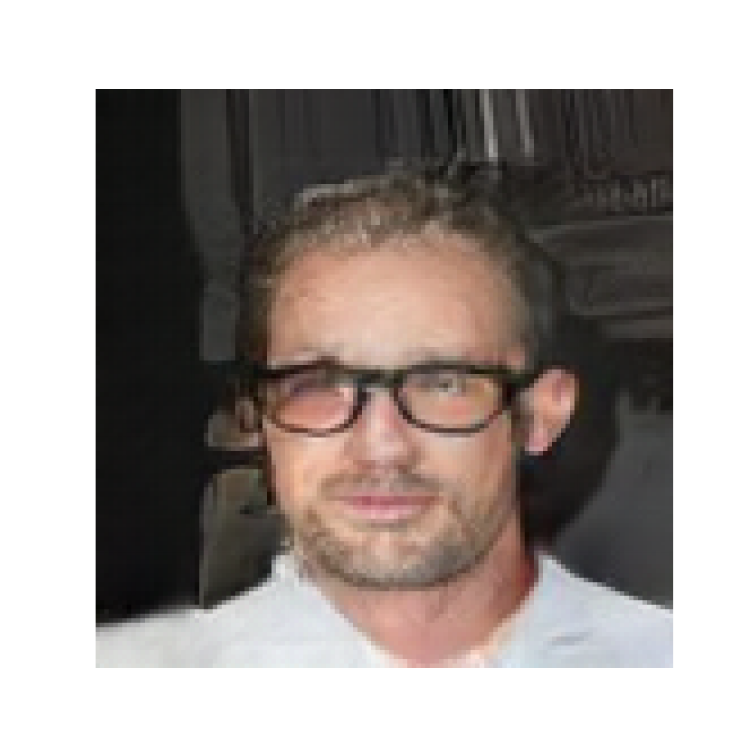} &
    \includegraphics[width = 1.4in]{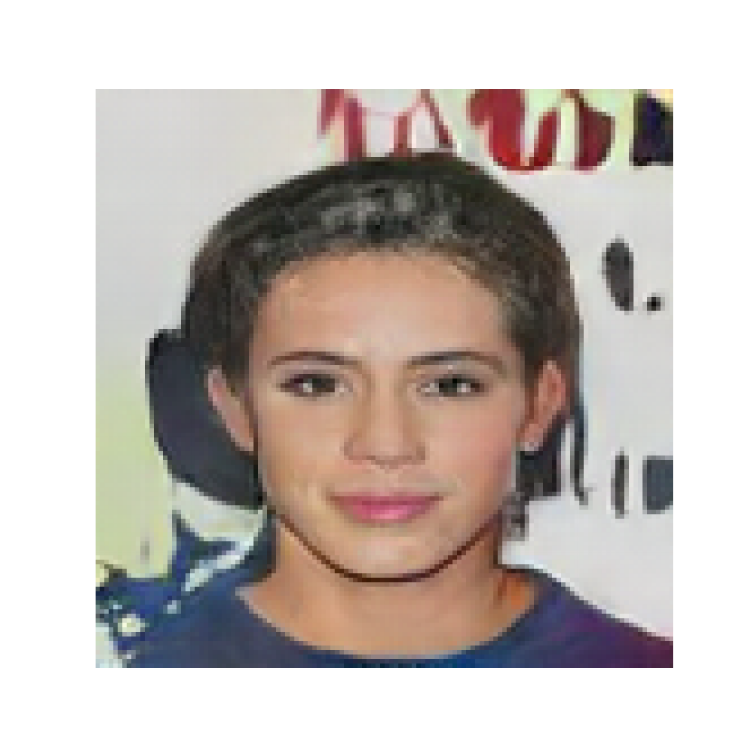} &
    \includegraphics[width = 1.4in]{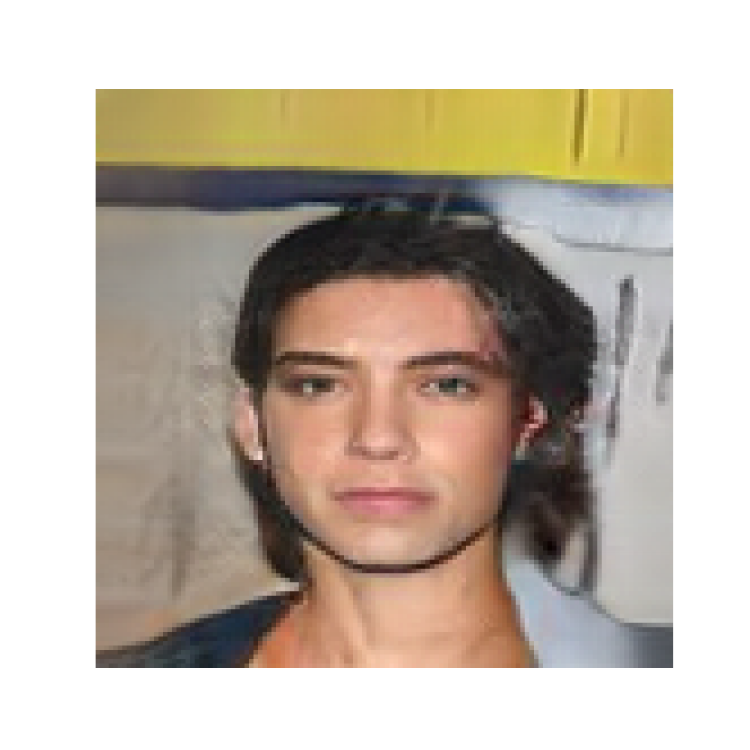} &
    \includegraphics[width = 1.4in]{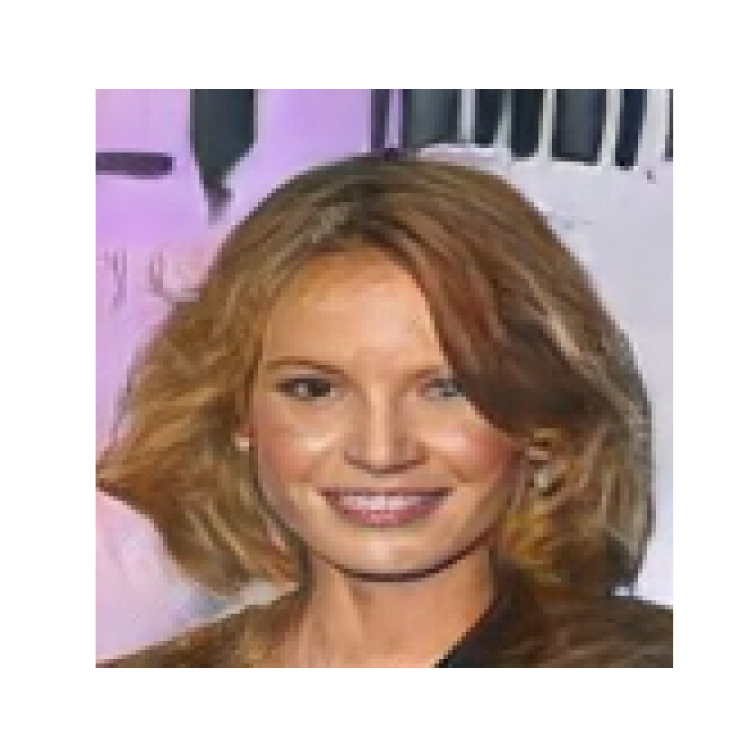} 
    
    \\
    \includegraphics[width = 1.4in]{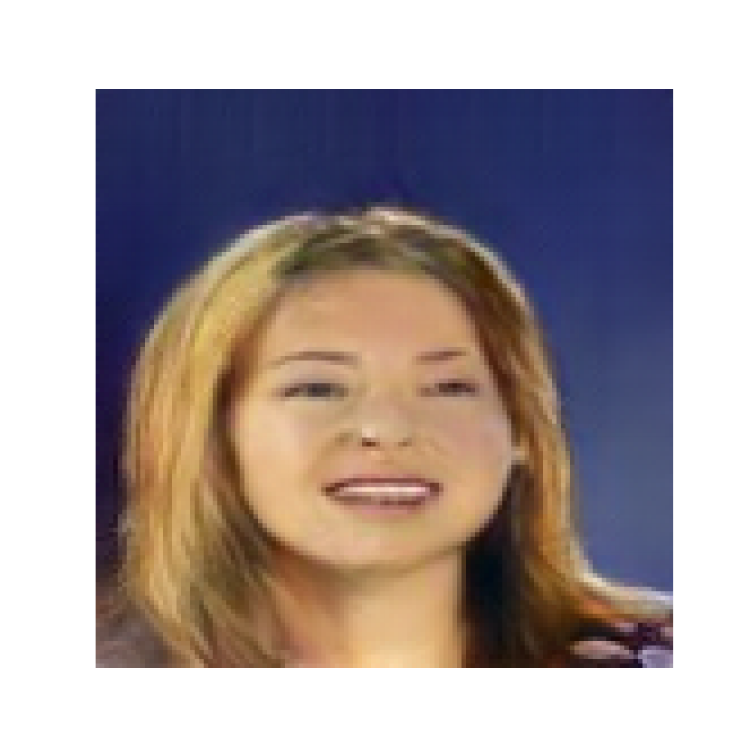} &
    \includegraphics[width = 1.4in]{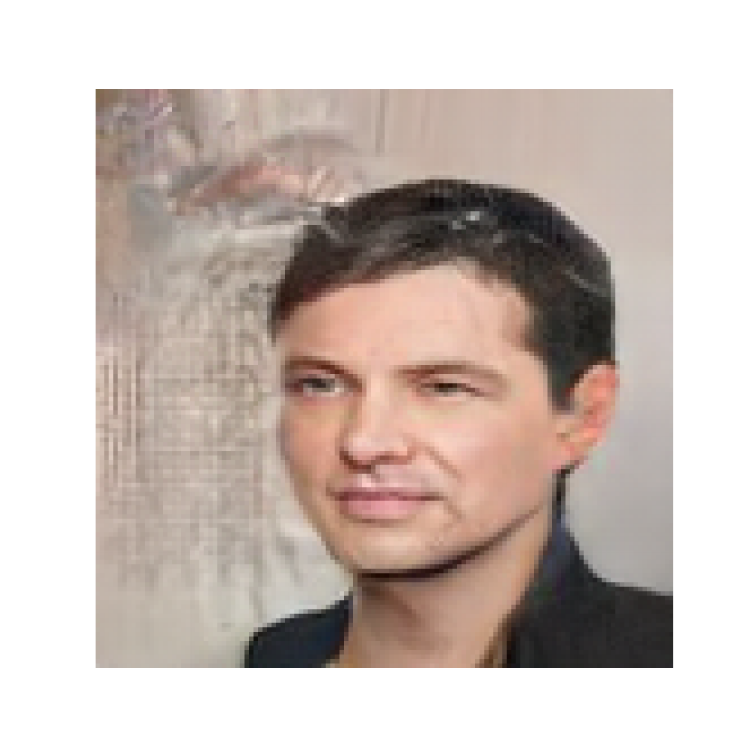} &
    \includegraphics[width = 1.4in]{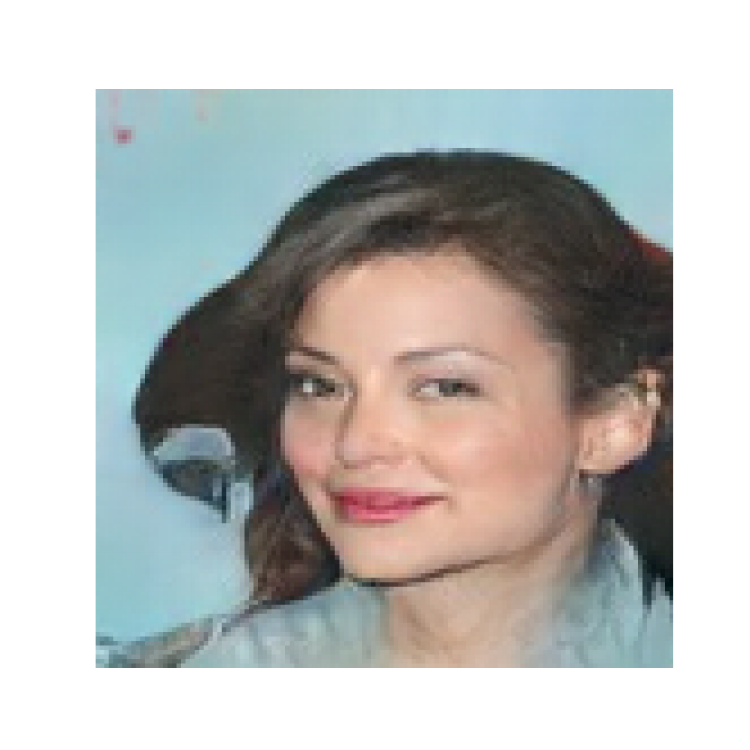} &
    \includegraphics[width = 1.4in]{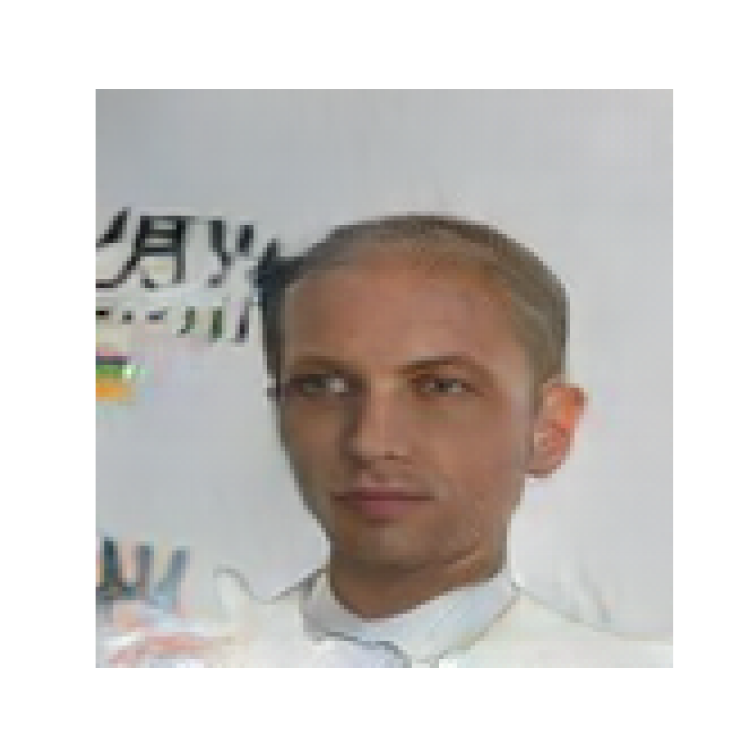} \\
    \end{tabular}
    \caption{acGAN - 3 discriminators}
    \end{subfigure}%
    
    \begin{subfigure}[H]{\textwidth}
    \centering
    \begin{tabular}{cccc}
    \includegraphics[width = 1.4in]{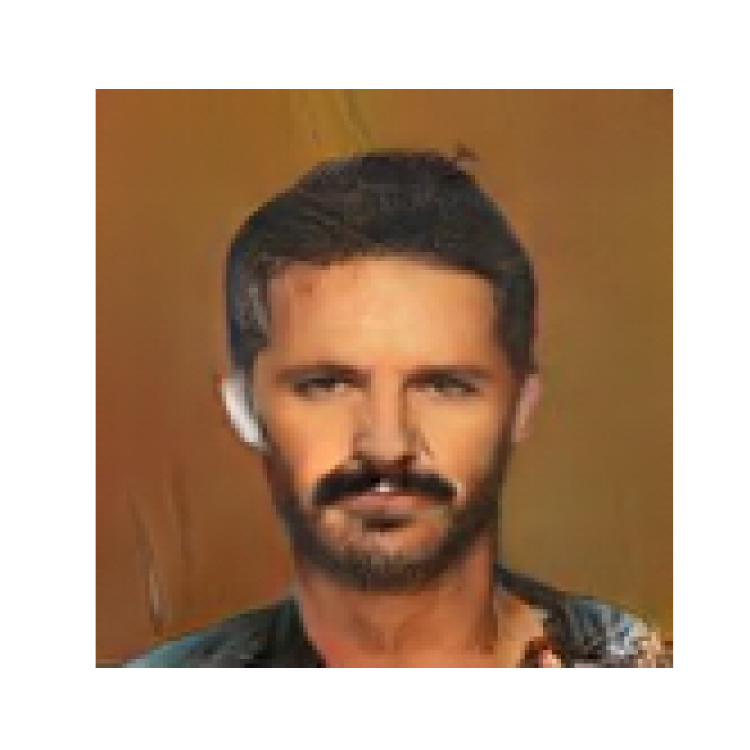}&
    \includegraphics[width = 1.4in]{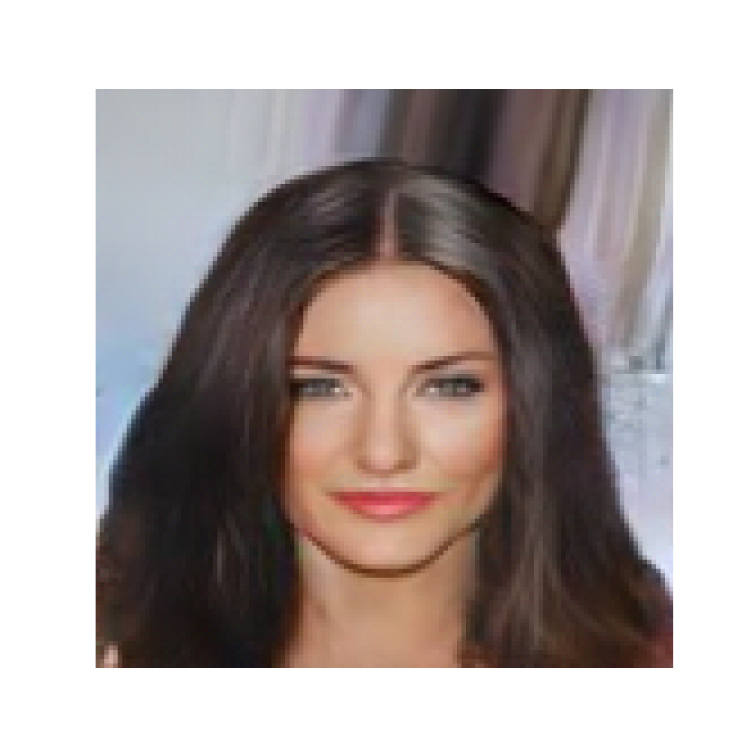} &
    \includegraphics[width = 1.4in]{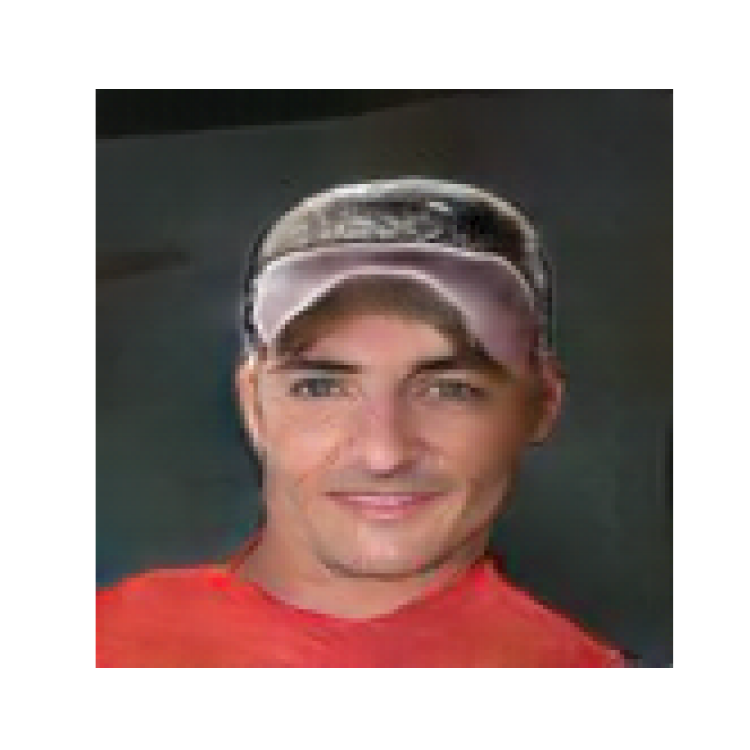} &
    \includegraphics[width = 1.4in]{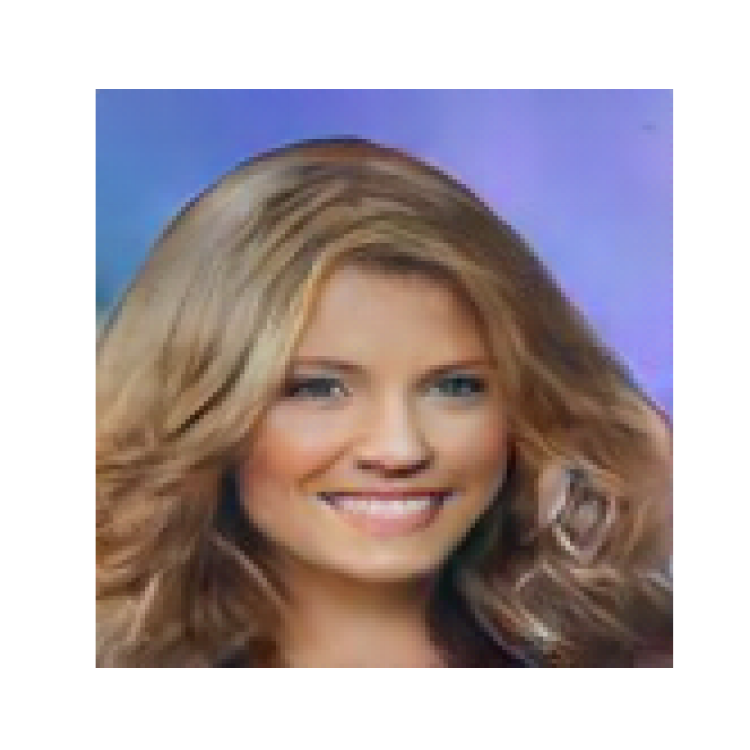} \\
    \includegraphics[width = 1.4in]{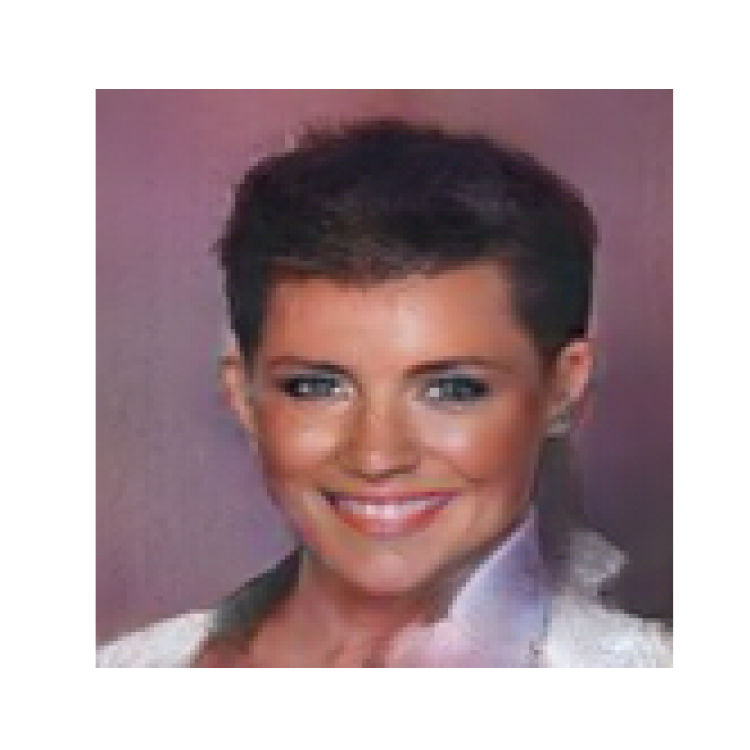} &
    \includegraphics[width = 1.4in]{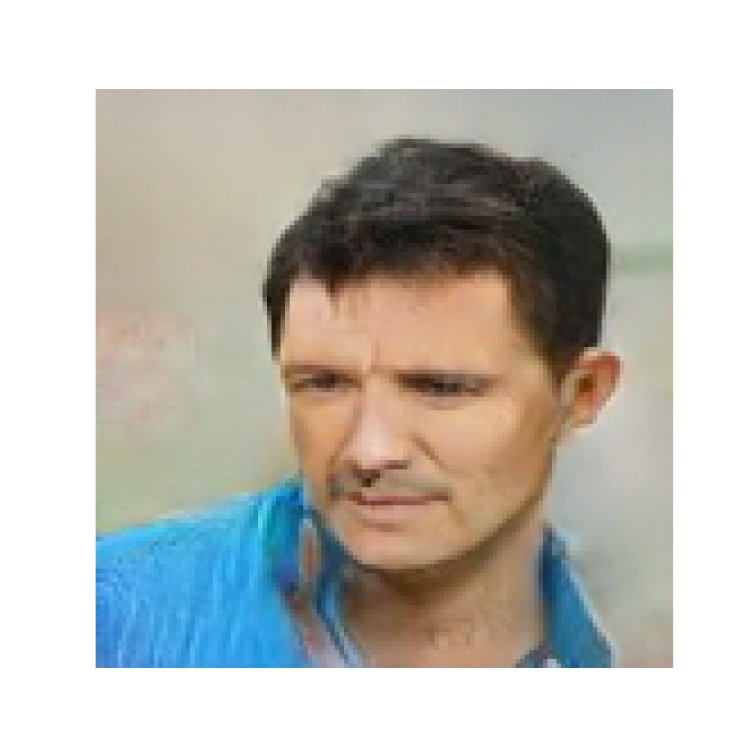} &
    \includegraphics[width = 1.4in]{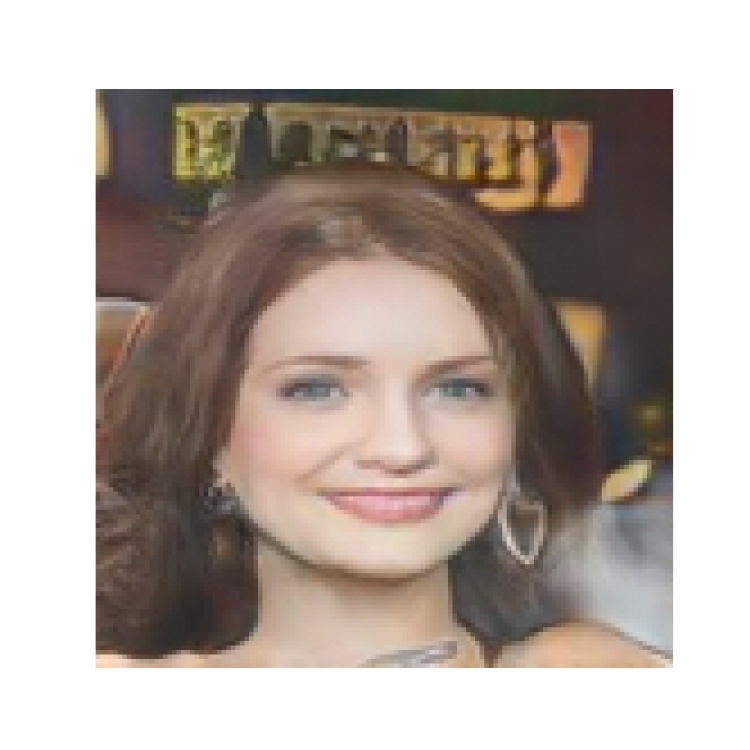} &
    \includegraphics[width = 1.4in]{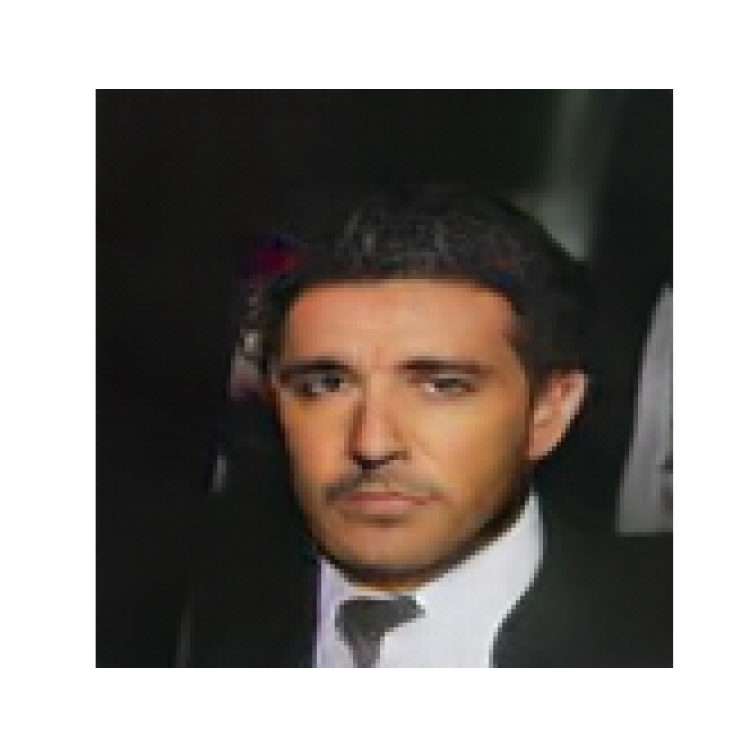} \\
    \includegraphics[width = 1.4in]{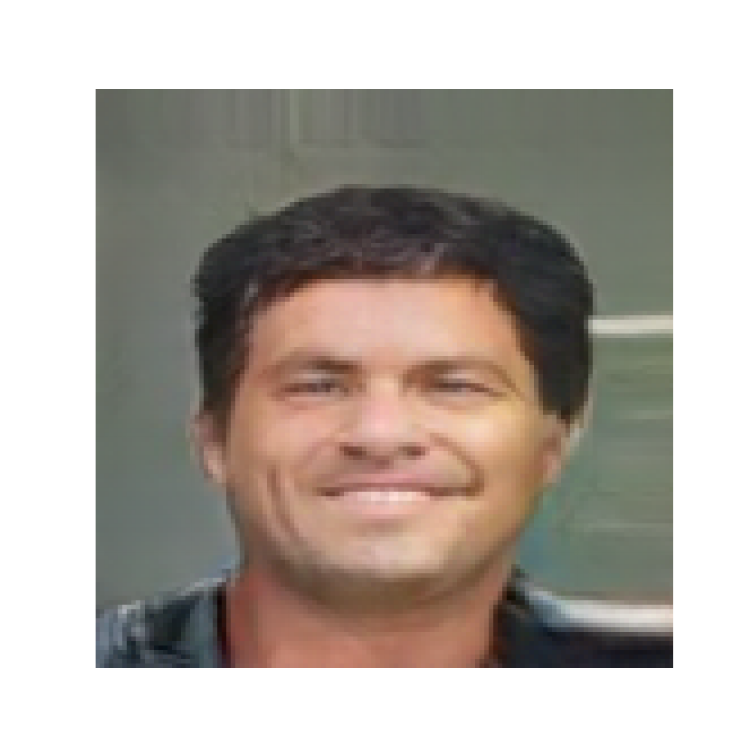} &
    \includegraphics[width = 1.4in]{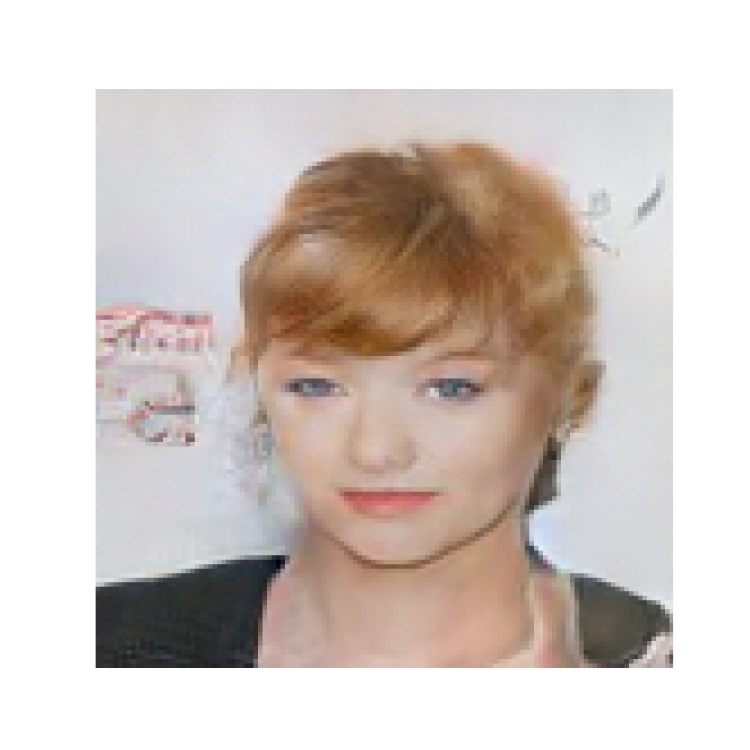} &
    \includegraphics[width = 1.4in]{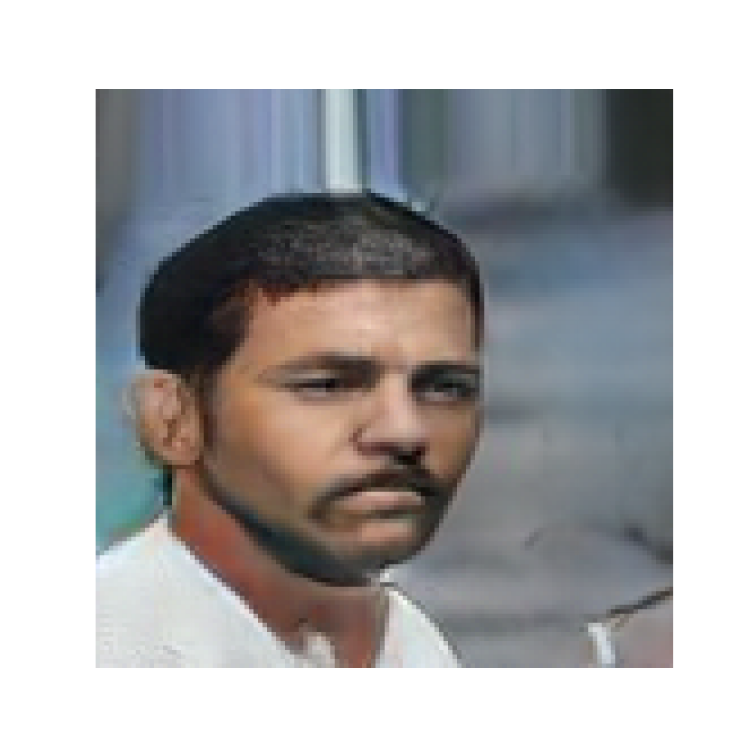} &
    \includegraphics[width = 1.4in]{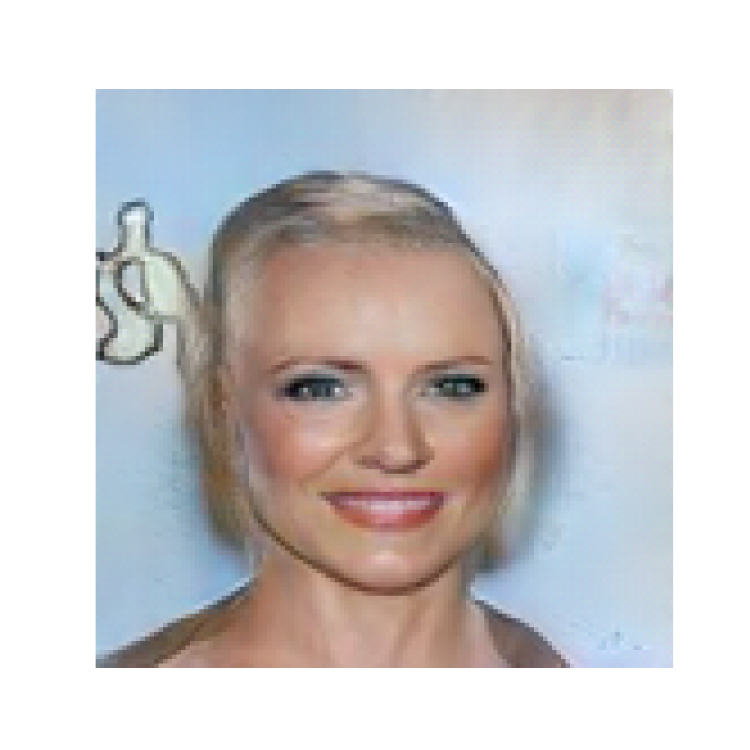} \\
    \end{tabular}
    \caption{acGAN - 5 discriminators}
    \end{subfigure}%
  \caption{128x128 CelebA samples for acGAN trained for 50 epochs with 3 and 5 discriminators.} 
  \label{fig:celeba_128}
\end{figure}

\end{document}